\documentclass[acmlarge]{acmart}

\AtBeginDocument{%
  }

\setcopyright{acmlicensed}
\copyrightyear{2018}
\acmYear{2018}
\acmDOI{XXXXXXX.XXXXXXX}

\acmJournal{POMACS}
\acmVolume{37}
\acmNumber{4}
\acmArticle{111}
\acmMonth{8}

\usepackage{booktabs}
\usepackage{amssymb} 
\usepackage{tabularx}
\usepackage{booktabs}
\usepackage{array}
\usepackage{ragged2e}
\usepackage{enumitem}
\usepackage{multirow}

\begin{document}

\title{Cyberbullying Governance on Social Media: A Unified Framework from Content Identification to Intervention}






\author{Yiting Huang}
\email{huangyiting@bupt.edu.cn}
\affiliation{%
  \institution{School of Cyberspace Security, Beijing University of Posts and Telecommunications}
  \city{Beijing}
  \country{China}
}

\author{Wenting Zhu}
\email{zwt@bupt.edu.cn}
\affiliation{%
  \institution{School of Cyberspace Security, Beijing University of Posts and Telecommunications}
  \city{Beijing}
  \country{China}
}

\author{Zekun Wang}
\email{ikun@bupt.edu.cn}
\affiliation{%
  \institution{School of Cyberspace Security, Beijing University of Posts and Telecommunications}
  \city{Beijing}
  \country{China}
}

\author{Qingpo Yang}
\email{2023111098@bupt.edu.cn}
\affiliation{%
  \institution{School of Cyberspace Security, Beijing University of Posts and Telecommunications}
  \city{Beijing}
  \country{China}
}

\author{Yakai Chen}
\email{chenyk@bupt.edu.cn}
\affiliation{%
  \institution{School of Cyberspace Security, Beijing University of Posts and Telecommunications}
  \city{Beijing}
  \country{China}
}

\author{Zihui Xu}
\email{2024111056@bupt.edu.cn}
\affiliation{%
  \institution{School of Cyberspace Security, Beijing University of Posts and Telecommunications}
  \city{Beijing}
  \country{China}
}

\author{Yueyue Zhang}
\email{zhangyueyue@bupt.edu.cn}
\affiliation{%
  \institution{School of Cyberspace Security, Beijing University of Posts and Telecommunications}
  \city{Beijing}
  \country{China}
}

\author{Sanchuan Guo}
\email{guosc@bupt.edu.cn}
\affiliation{%
  \institution{School of Cyberspace Security, Beijing University of Posts and Telecommunications}
  \city{Beijing}
  \country{China}
}

\author{Xi Zhang}
\authornote{Corresponding author.}
\email{zhangx@bupt.edu.cn}
\affiliation{%
  \institution{School of Cyberspace Security, Beijing University of Posts and Telecommunications}
  \city{Beijing}
  \country{China}
}

\thanks{This work was supported by the National Key Research and Development Program of China (No.xxx), and the Key Laboratory of Trustworthy Distributed Computing and Service (MOE)}



\begin{abstract}
The proliferation of social media platforms and online communities has inadvertently catalyzed the spread of cyberbullying, hate speech, and other forms of online toxicity, making the effective governance of such harm a critical societal and computational challenge.
While significant strides have been made in automating content moderation, existing research predominantly treats cyberbullying governance as passive, isolated detection at the post level. 
This reductionist view overlooks the continuous behavioral dynamics of users, the structural diffusion of toxic events, and the critical need for proactive mitigation.
To bridge these gaps, this paper proposes a unified full-lifecycle governance framework that shifts the paradigm of cyberbullying governance from isolated static detection toward integrated, continuous, and proactive moderation.
Drawing on cyberbullying research and adjacent fields, we systematically synthesize the state-of-the-art literature across four interconnected stages: 
(1) Content Identification, 
(2) User and Behavior Modeling,
(3) Diffusion Dynamics and Early Warning, 
and (4) Intervention and Governance.
Furthermore, we review available datasets and evaluation practices, and discuss emerging challenges including multimodality, explainability, algorithmic fairness, and the dual-use risks of generative AI, providing a roadmap for future research toward a safer and more resilient digital ecosystem.
\end{abstract}



\received{20 February 2007}
\received[revised]{12 March 2009}
\received[accepted]{5 June 2009}

\maketitle


\section{Introduction}\label{sec:intro}

Social media platforms, instant messaging services, and online communities have transformed human communication by enabling large-scale connectivity, information sharing, and public expression \cite{chan2021cyberbullying}. 
At the same time, they have also amplified cyberbullying, hate speech, abusive language, and other forms of online toxicity \cite{mishra2019tackling}. 
Unlike offline bullying, cyberbullying is not constrained by physical proximity or time. 
Enabled by anonymity, persistence, networked visibility, and algorithmic amplification, isolated hostile expressions can rapidly escalate into coordinated harassment and wider social harm \cite{thomas2021sok}. 
Cyberbullying governance is therefore not only a problem of harmful content moderation, but also a broader challenge for online safety, platform governance, and digital public life.

A central difficulty is that cyberbullying does not have stable boundaries in the literature. Prior reviews have shown substantial inconsistency in how it is defined and measured across disciplines \cite{ray2024cyberbullying}. 
While classic bullying research emphasizes repetition, hostile intent, and power imbalance, many computational studies operationalize cyberbullying as offensive or toxic content at the post level. 
This mismatch is consequential because cyberbullying overlaps with hate speech, harassment, abusive language, and trolling, yet is not reducible to any one of them \cite{banko2020unified}. 
Harm may be explicit or implicit, individual or coordinated, and textual or multimodal \cite{vora2023multimodal}. 
A governance-oriented analysis therefore needs to preserve the specificity of cyberbullying while situating it within the broader landscape of online abuse.

More importantly, cyberbullying is rarely an isolated event. 
It often unfolds as a dynamic process spanning emergence, escalation, diffusion, and response \cite{yi2023session,thomas2021sok}. 
Harmful interactions may begin with subtle hostility, intensify through repeated exchanges or role asymmetries, spread through bystander participation and platform visibility, and eventually trigger moderation, counterspeech, or other governance responses \cite{kao2019understanding}.  
From this perspective, cyberbullying should be understood not merely as harmful content to be detected, but as a full-lifecycle governance problem that evolves over time and across actors, interactions, and platforms \cite{milosevic2022artificial}.

Recent technical research has made substantial progress in harmful content detection, moving from keyword and feature-based methods toward contextual language models, session-based modeling, and multimodal moderation \cite{vora2023multimodal,Yi2023LearningLH}. 
At the same time, adjacent work has expanded the technical scope of online harm governance toward user and behavior modeling, diffusion analysis, counterspeech, and platform moderation pipelines \cite{singhal2023sok,bonaldi2024nlp}. 
These advances reveal a deeper point: effective cyberbullying governance is not a single-task prediction problem. 
It is a multi-stage process that spans the identification of harmful content, the modeling of actors and interaction patterns, the anticipation of escalation and diffusion, and the design of interventions that can reduce harm without imposing disproportionate censorship.

Motivated by this view, this paper organizes cyberbullying governance into four interconnected stages: content identification, user and behavior modeling, propagation and scale prediction, and intervention and governance. 
The first stage concerns recognizing harmful signals across text, images, video, and conversational context \cite{salawu2017approaches}. 
The second focuses on social roles, relational structure, coordinated attackers, and malicious or automated accounts \cite{liu2023creativity,alfurayj2024exploring}. 
The third examines escalation, cascades, topic evolution, and early warning. 
The fourth addresses how platforms and automated systems respond through moderation, counterspeech, policy enforcement, and broader governance mechanisms \cite{chung2024effectiveness}. 
Together, these stages provide a full-lifecycle view of cyberbullying governance, capturing how harmful interactions emerge, spread, and are addressed on social media.

Existing work provides valuable foundations, but remain fragmented across tasks and communities. 
Recent studies often emphasize cyberbullying detection, multimodal identification, counterspeech and intervention, or broader moderation and large-model governance \cite{dai2023review,alfurayj2024exploring,li2021capturing,chen2025comprehensive}. 
As summarized in Table \ref{tab:comparison}, few efforts connect these strands into a unified view spanning content identification, behavioral modeling, diffusion dynamics, and intervention mechanisms. 
This fragmentation makes it difficult to understand cyberbullying as an evolving governance problem rather than an isolated detection task.

\begin{table*}[t]
\centering
\caption{Comparison with representative surveys in the domain of cyberbullying governance. Symbols: $\checkmark$ = Dedicated section-level coverage; $\sim$ = Partial mention; $-$ = Not covered.}
\label{tab:comparison}
\resizebox{\textwidth}{!}{%
\begin{tabular}{llcccccl}
\toprule
\textbf{Work} & \textbf{Core Focus} & \textbf{Content} & \textbf{Behavior} & \textbf{Diffusion \&} & \textbf{Proactive} & \textbf{Full} & \textbf{Key Differences} \\
& & \textbf{ID} & \textbf{Modeling} & \textbf{Warning} & \textbf{Intervention} & \textbf{Lifecycle} & \\
\midrule

Dai et al. \citep{dai2023review} 
& Cyberbullying detection and governance strategies 
& $\checkmark$ & $\sim$ & $-$ & $\sim$ & $-$ 
& Detection-oriented \\

Woo et al. \citep{woo2023cyberbullying} 
& Conceptualization, characterization, and detection 
& $\checkmark$ & $\sim$ & $-$ & $-$ & $-$ 
& Limited intervention \\

V et al. \citep{sahanav2023asl} 
& Taxonomy, datasets, and detection approaches 
& $\checkmark$ & $\sim$ & $-$ & $\sim$ & $-$ 
& No lifecycle view \\

Vora et al. \citep{vora2023multimodal} 
& Multimodal cyberbullying detection 
& $\checkmark$ & $-$ & $-$ & $-$ & $-$ 
& Multimodal only \\

Alfurayj et al. \citep{alfurayj2024exploring} 
& Bystander contagion in cyberbullying 
& $\sim$ & $\checkmark$ & $\sim$ & $-$ & $-$ 
& Role-focused \\

Li et al. \citep{li2021capturing} 
& Information diffusion in social networks 
& $-$ & $\sim$ & $\checkmark$ & $-$ & $-$ 
& Diffusion-focused \\

Bonaldi et al. \citep{bonaldi2024nlp} 
& Counterspeech and harm mitigation 
& $-$ & $-$ & $-$ & $\checkmark$ & $-$ 
& Intervention only \\

Hee et al. \citep{hee2024recent} 
& Hate speech moderation with large models 
& $\checkmark$ & $-$ & $-$ & $\checkmark$ & $-$ 
& Moderation-focused \\

Chen et al. \citep{chen2025comprehensive} 
& LLM-based content moderation 
& $\checkmark$ & $-$ & $-$ & $\checkmark$ & $-$ 
& LLM moderation only \\

\midrule
\textbf{Our work} 
& \textbf{Full-lifecycle cyberbullying governance} 
& $\checkmark$ & $\checkmark$ & $\checkmark$ & $\checkmark$ & $\checkmark$ 
& \textbf{Unified lifecycle view} \\

\bottomrule
\end{tabular}%
}
\end{table*}


To address these gaps, this paper proposes a unified full-lifecycle framework for cyberbullying governance on social media. 
Rather than treating cyberbullying as an isolated harmful-content detection task, we organize the literature into four interconnected stages:
(1) Content Identification, (2) User and Behavior Modeling, (3) Diffusion Dynamics and Early Warning, and (4) Intervention and Governance. 
While cyberbullying remains the core focus, we also draw on adjacent work in hate speech, abusive language, and online harassment where it helps clarify conceptual boundaries or provides transferable technical insights.


Compared with existing work, this paper offers the following main contributions:
\begin{itemize}
    \item \textbf{A unified full-lifecycle governance framework.} We propose a four-stage framework that captures cyberbullying as a dynamic governance problem spanning content identification, user and behavior modeling, diffusion dynamics and early warning, and intervention and governance, bridging research communities that are often treated in isolation.
    \item \textbf{A cross-stage synthesis of LLMs in cyberbullying governance.} We systematically analyze how LLMs are reshaping each governance stage—from implicit detection and behavior simulation, to diffusion prediction and opinion modeling, to counter-narrative generation and multi-agent intervention evaluation—while examining their dual-use risks in adversarial settings.
    \item \textbf{A structured analysis of datasets and open challenges.} We examine 21 publicly available datasets and discuss open challenges including lifecycle-aware modeling, explainability and accountability, algorithmic bias and low-resource governance, and the adversarial risks of large language models.
\end{itemize}

The remainder of this paper is organized as follows. 
Section \ref{sec:basics} introduces the basic concepts, taxonomies, and lifecycle of cyberbullying. 
Section \ref{sec:content_id} examines content identification methods.
Section \ref{sec:behavior} examines user and behavior modeling.
Section \ref{sec:propagation} examines diffusion dynamics and early-warning methods. 
Section \ref{sec:intervention} discusses intervention and governance strategies.
Section \ref{sec:datasets} summarizes datasets, benchmarks, and evaluation protocols. 
Section \ref{sec:challenges} outlines major challenges and future directions.
Section \ref{sec:conclusion} concludes the paper.

\section{Basics and Taxonomy}\label{sec:basics} 
\subsection{Task Definitions and Distinctions}\label{subsec:definition}
With the proliferation of social media platforms, harmful and aggressive content has manifested in diverse forms online. 
Among them, cyberbullying has attracted increasing attention as a severe form of online harm. 
However, in existing literature, cyberbullying is often conflated with related concepts such as hate speech, toxic language, and offensive language, leading to ambiguity in task formulation and modeling objectives. 
In particular, many studies in natural language processing treat these tasks uniformly as text classification problems, overlooking their fundamental differences in terms of target, behavioral structure, and temporal dynamics. 
Therefore, a systematic conceptual distinction is necessary.

\begin{itemize}
    \item \textbf{Cyberbullying} is commonly regarded as a form of bullying conducted through online media. Bullying itself is typically defined as repeated and deliberate aggressive behavior by an individual or group toward a person who is in a relatively vulnerable position \cite{Yi2022SessionbasedCD}. Although specific formulations vary across studies, two characteristics are consistently emphasized: repetition and power imbalance, both of which are considered key indicators in identifying cyberbullying behavior.
    \item \textbf{Hate speech} generally refers to expressions that attack, demean, or incite violence or hatred against a group based on protected characteristics, such as appearance, religion, ancestry, ethnicity, sexual orientation, or gender identity \cite{HateSpeeech}. Such expressions may appear in various linguistic forms, including subtle or even humorous formulations, but are fundamentally characterized by identity-based targeting.
    \item \textbf{Toxic language} is often defined in a broader sense as language that is rude, disrespectful, or unreasonable, and is likely to disrupt conversations or drive participants away \cite{toxic}, emphasizing the negative impact on the conversational environment.
    \item \textbf{Offensive language} refers to expressions that may insult or hurt others’ feelings, including profanity, derogatory remarks, or vulgar language \cite{offensive}. Its identification often depends on both lexical usage and contextual interpretation.
\end{itemize}


Although these concepts overlap in practice, prior studies generally identify repetition as a key property that distinguishes cyberbullying from other abusive language tasks. 
Hate speech is primarily directed toward specific groups and is often associated with bias or discrimination, yet is typically identified based on individual instances without requiring temporal persistence. 
Similarly, toxic and offensive language are commonly modeled as properties of single utterances, focusing on linguistic aggressiveness or conversational disruption rather than temporal patterns. 
In contrast, cyberbullying is understood as a behavior that unfolds over time, where identification requires not only the analysis of individual expressions but also the continuity of interactions—suggesting that single-utterance analysis is insufficient to capture its nature.


Building upon these defining properties, prior work further delineates these concepts across three key dimensions. 
First, regarding the target of abuse, cyberbullying typically singles out a specific individual victim, whereas hate speech is systematically directed at groups defined by protected characteristics, and toxic and offensive language often occur without a clearly specified target. 
Second, in terms of temporal scope, cyberbullying inherently spans multiple, sustained interactions, while the remaining three are predominantly formulated as instance-level classification tasks. 
Finally, these phenomena capture fundamentally different vectors of harm: toxic and offensive language primarily disrupt online civility, hate speech perpetuates systemic identity-based hostility, and cyberbullying inflicts sustained, targeted psychological damage. 
These fundamental differences underscore why effective cyberbullying governance demands a continuous, lifecycle-aware framework rather than a static text classification approach.


\subsection{Lifecycle of Cyberbullying}\label{subsec:lifecycle}

\begin{figure}[t]
  \includegraphics[width=\columnwidth]{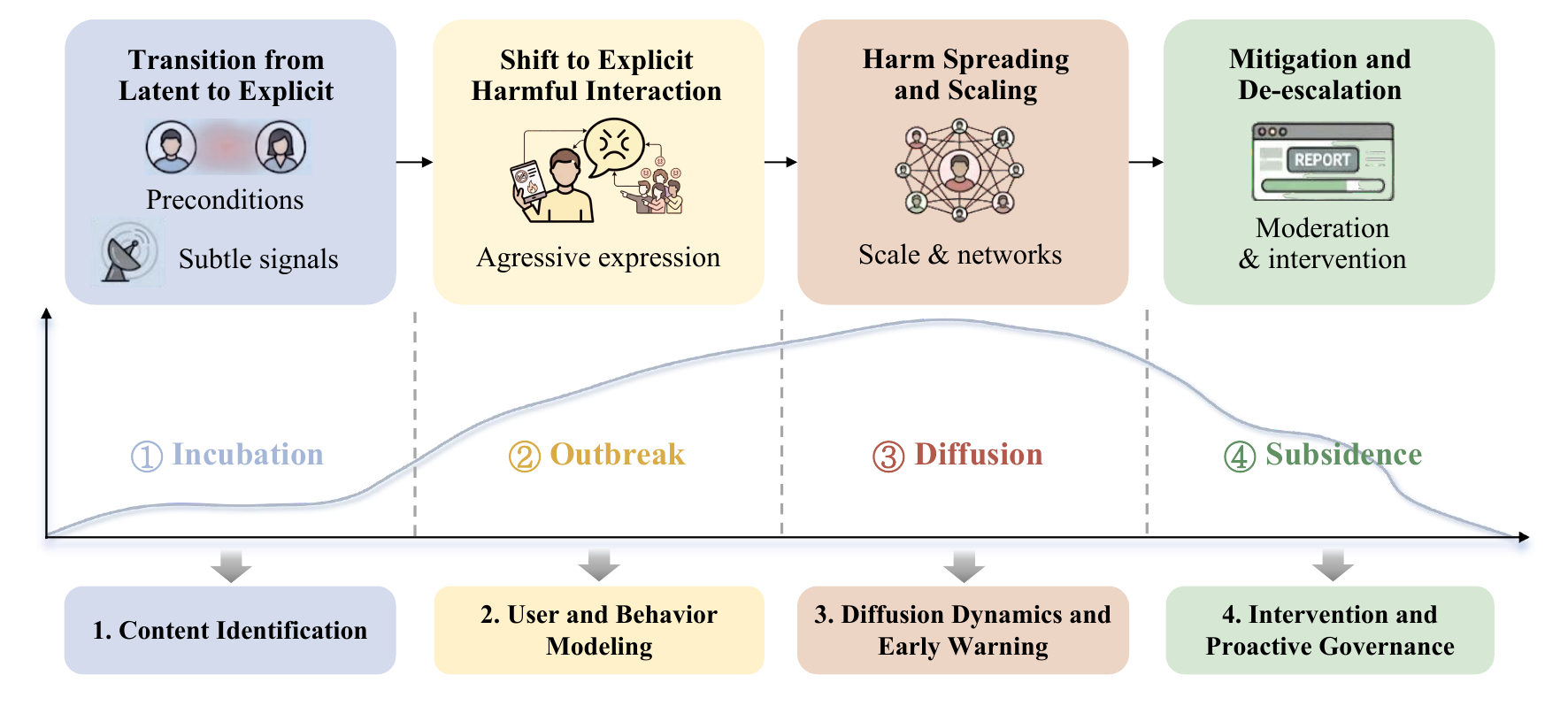}
  \caption{The four-stage lifecycle of cyberbullying and online toxicity.}
  \label{fig:lifecycle}
\end{figure}


Building on the distinctions above, we adopt a four-stage lifecycle view to organize both the phenomenon and the corresponding governance tasks: \textit{incubation}, \textit{outbreak}, \textit{diffusion}, and \textit{subsidence}, as illustrated in Figure~\ref{fig:lifecycle}. 
These stages should not be interpreted as rigidly separated or universally sequential. 
Rather, they serve as an abstract framework for describing how online harm may emerge, intensify, spread, and eventually be addressed in social media environments. 
Crucially, this process forms a continuous loop, where the outcomes of mitigation strategies feed back into the ecosystem, shaping future risks and subsequent interactions.

\textbf{Incubation.} 
In the incubation stage, harmful dynamics are not yet fully visible as explicit cyberbullying incidents, but risk factors and preconditions may already be present. 
These may include antagonistic interaction histories, vulnerable target positions, prior role asymmetries, or community norms that tolerate hostility. 
In many cases, early signals are subtle, such as repeated negative cues, sarcasm, exclusionary behavior, or the gradual concentration of hostile attention around a target. 
From a governance perspective, this stage—characterized by the transition from latent to explicit harm—is closely related to contextual monitoring, user and role modeling, and early-warning mechanisms.

\textbf{Outbreak.} 
The outbreak stage refers to the transition from latent hostility to explicit harmful interaction. 
At this point, aggressive or abusive expressions become observable, often through insults, threats, humiliation, or repeated antagonistic exchanges directed toward a target. 
Unlike general offensive language, cyberbullying at this stage is often characterized not only by the harmful content itself, but also by the recurrence and targeting pattern of the interaction. 
Detection tasks at this stage therefore require more than identifying isolated toxic utterances; they also benefit from session-level context, role information, and conversational structure.

\textbf{Diffusion.} 
Once cyberbullying emerges, it may spread beyond the initial interaction through network effects, leading to harm spreading and scaling. 
Harm can intensify through bystander participation, reposting, recommendation mechanisms, coordinated attacks, or migration across threads and platforms. 
In this stage, the scale and impact of cyberbullying are shaped not only by the original aggressor, but also by visibility, attention dynamics, and collective behavior. This makes diffusion analysis, escalation modeling, and scale prediction particularly relevant. 
From a governance perspective, the diffusion stage is where timely intervention becomes crucial, since delayed responses may allow localized hostility to develop into broader harassment.

\textbf{Subsidence.} 
The subsidence stage concerns how cyberbullying is mitigated, de-escalated, or brought under control. 
This may occur through platform moderation, reporting and review processes, counterspeech, community intervention, or changes in attention and participation patterns. 
Importantly, mitigation and de-escalation do not necessarily imply complete resolution. 
Harm may persist through psychological effects on the target, reputational damage, or continued circulation of abusive content in other spaces. 
For this reason, governance at this stage includes not only removal or suppression, but also response design, accountability mechanisms, and support-oriented interventions.
Ultimately, the outcomes of these interventions act as a feedback loop, reshaping community norms and establishing the preconditions for the next cycle of incubation.

Taken together, this lifecycle view highlights that cyberbullying governance is not confined to post hoc detection. 
Different governance tasks become salient at different stages, from identifying early signals and modeling risky interaction patterns, to forecasting escalation and designing timely interventions. 
This perspective also motivates the broader taxonomy of governance technologies introduced below.

\subsection{Taxonomy of Governance Technologies}\label{subsec:tech_taxonomy}

\begin{figure}[t]
  \includegraphics[width=\columnwidth]{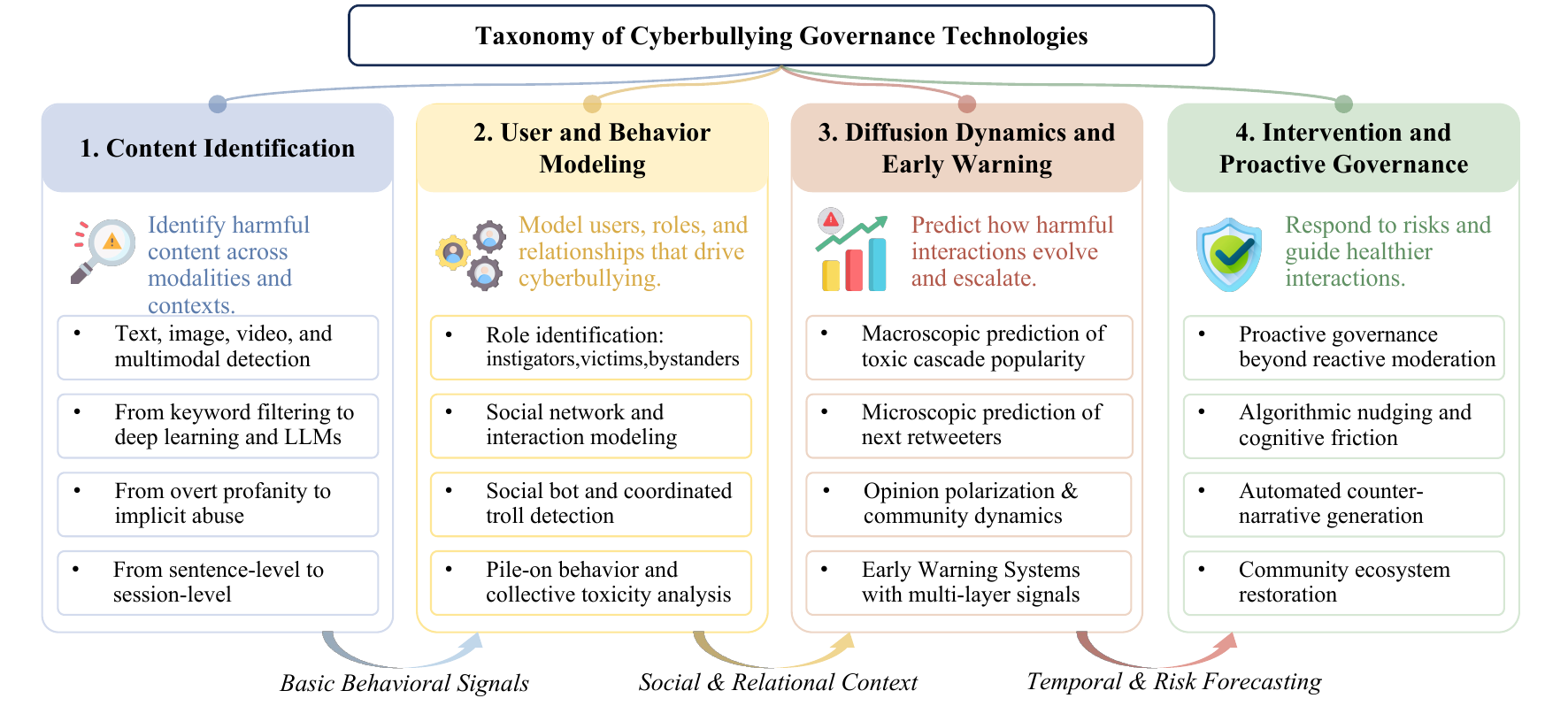}
  \caption{The proposed closed-loop taxonomy of cyberbullying governance technologies.}
  \label{fig:taxonomy}
\end{figure}

We organize cyberbullying governance technologies into a function-oriented taxonomy comprising four interconnected categories: (1) Content Identification, (2) User and Behavior Modeling, (3) Diffusion Dynamics and Early Warning, and (4) Intervention and Proactive Governance, as illustrated in Figure~\ref{fig:taxonomy}. 
Compared with taxonomies that focus solely on algorithmic techniques or data modalities, this functional framework better captures how different methodologies sequentially and synergistically contribute to mitigating cyberbullying across its full lifecycle~\cite{singhal2023sok,milosevic2022artificial}.

\textbf{Content identification.} 
This category represents the foundational layer of governance, focusing on the recognition of harmful signals across various modalities, including text, images, and video. 
As detailed in Section \ref{sec:content_id}, the field has evolved from dictionary-based keyword filtering to context-aware deep learning and Large Language Models (LLMs) \cite{vora2023multimodal}. 
This progression has enabled the detection of increasingly sophisticated harm, shifting the focus from overt profanity to implicit abuse \cite{yi2023session}, such as sarcasm and microaggressions, and from isolated sentence-level analysis to session-level contextual awareness.

\textbf{User and behavior modeling.} 
To move beyond isolated utterances, this category examines the social actors and relational structures that drive cyberbullying \cite{alfurayj2024exploring}. 
As explored in Section \ref{sec:behavior}, this involves identifying participant roles (e.g., instigators, victims, and various types of bystanders), modeling social network dynamics through graph-based methods, and detecting social bots or coordinated troll accounts that amplify toxicity. 
These technologies are critical for identifying pile-on behaviors where the harm stems from the collective structure of interactions rather than the content of a single post.

\textbf{Diffusion dynamics and early warning.} 
This category introduces temporal and risk-aware reasoning by anticipating how harmful interactions evolve and spread through social systems \cite{li2021capturing}. 
Section \ref{sec:propagation} reviews methodologies for predicting the macroscopic popularity of toxic cascades, identifying the microscopic path of next-retweeters, and analyzing opinion polarization. 
These models culminate in Early Warning Systems (EWS) that synthesize multi-layer signals—content, propagation, and behavior—to provide actionable risk assessments before a localized incident escalates into large-scale harassment.

\begin{figure*}[!t]
\centering
\includegraphics[width=\textwidth]{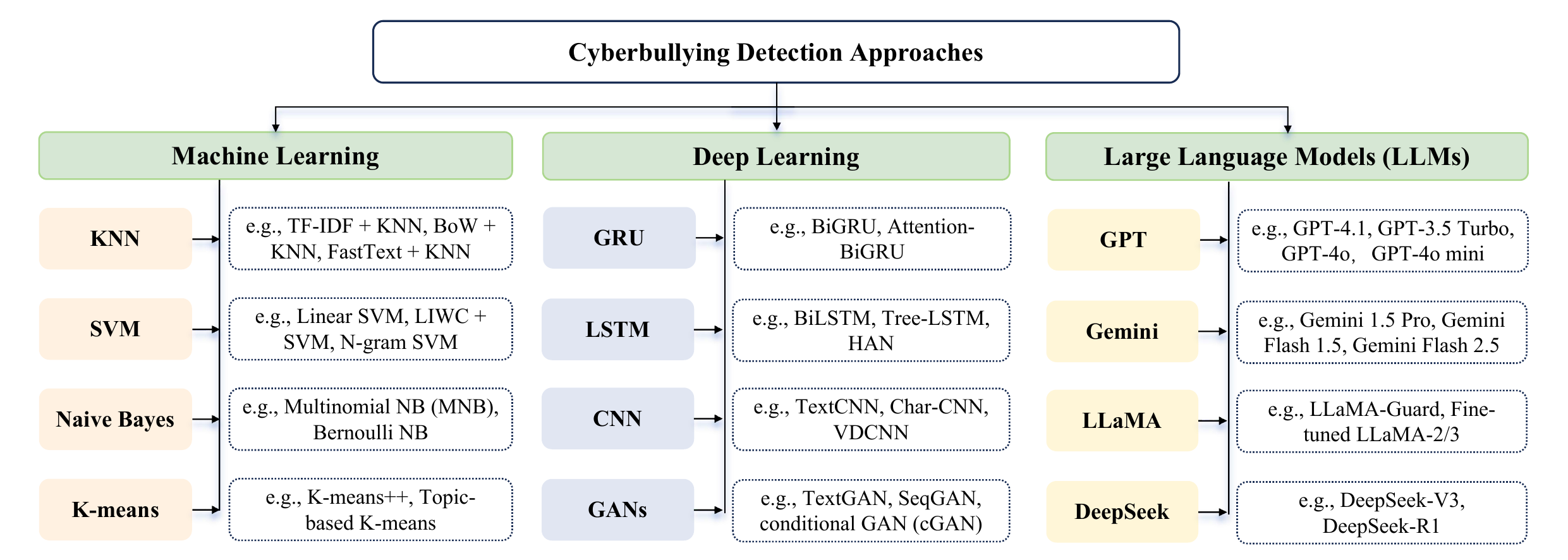}
\caption{Taxonomy of Cyberbullying Detection Approaches.}
\label{fig:cyberbully-detect}
\end{figure*}

\textbf{Intervention and proactive governance.} 
The final category covers the mechanisms through which platforms and automated systems respond to identified risks \cite{hee2024recent,bonaldi2024nlp}. 
Moving beyond passive, post-hoc measures like content deletion or account bans, Section \ref{sec:intervention} highlights a paradigm shift toward proactive governance. 
This includes intelligent nudging to introduce cognitive friction, and the automated generation of Counter-Narratives (CN) grounded in empathy and factual clarification. 
Such mechanisms aim not only to suppress harm but to restore the community ecosystem and guide user behavior toward self-correction.

These four categories are analytically distinct but tightly coupled, forming a resilient closed-loop ecosystem. 
Content identification provides the basic behavioral signals, while user and behavior modeling adds the necessary social and relational context. 
Diffusion dynamics introduces temporal and risk-aware forecasting, allowing intervention mechanisms to translate recognition and prediction into timely action. 
Ultimately, proactive interventions reshape user behaviors and network topologies, generating new data distributions that feed back into content identification, thereby driving the continuous evolution of the entire governance system.


\section{Content Identification}
\label{sec:content_id}

This section reviews the evolution of cyberbullying detection methods, spanning architectural advances from machine learning and deep learning to large language models, and breakthroughs across three key dimensions: from explicit to implicit cyberbullying, from sentence-level to session-level context, and from textual to multimodal data. 
A detailed classification diagram is shown in Figure~\ref{fig:cyberbully-detect}.

\subsection{Machine Learning Methods on Cyberbullying Detection}
Traditional machine learning methods lay the crucial foundational groundwork for the automated detection of cyberbullying. 
In the early stages of this research domain, machine learning algorithms effectively enhance the detection of cyberbullying and significantly improve accuracy by extracting and leveraging handcrafted features such as text examples, user behavioral patterns, and demographic characteristics \cite{kagi2025cyberbullying}. 
By training algorithms on historical, annotated datasets, these models acquire the capability to classify new, unseen cases, thereby drastically shortening manual investigation times and reducing human resource consumption \cite{desai2021cyber}. 

Machine learning methods in this domain span supervised, unsupervised, and semi-supervised paradigms. 
Supervised approaches are most widely adopted, with prominent examples including SVM, KNN, Naive Bayes, Logistic Regression, and ensemble methods such as Random Forests and Decision Trees~\cite{osisanwo2017supervised,bansal2021comparison,muneer2020comparative,thakur2020supervised}. 
Unsupervised methods mainly rely on clustering techniques such as K-means~\cite{naeem2023unsupervised}, while semi-supervised methods combine clustering with supervised algorithms to reduce annotation costs. 
Among these, SVM and Random Forests historically perform particularly well due to their robustness in handling high-dimensional, sparse data generated by text features.

However, despite their historical significance, traditional machine learning approaches inherently suffer from severe methodological bottlenecks, primarily stemming from their reliance on manual feature engineering. At this stage, feature extraction relies heavily on statistical word frequency techniques such as TF-IDF, the Bag-of-Words (BoW) model, and N-grams, supplemented by Part-of-Speech (POS) tagging and predefined sentiment lexicons. 
This shallow, discrete feature representation leads to significant dimensional limitations that restrict the evolution of early cyberbullying detection across the three key dimensions identified in this study:
\begin{itemize}
  \item \textbf{Explicit vs. Implicit Bullying:} Traditional machine learning methods rely heavily on abusive word dictionaries and syntactic matching, making them adept at capturing explicit cyberbullying containing overt profanity, slurs, and direct threats. However, they are highly prone to failure when faced with implicit cyberbullying—such as sarcasm, microaggressions, peer exclusion, or passive-aggressive comments—that do not contain inherently sensitive words. As shown by \citet{caselli2020feel}, the accuracy of vocabulary-matching classifiers drops sharply when dealing with covertly offensive language.Traditional models fail to grasp the pragmatic meaning behind polite but malicious phrasing.
  \item \textbf{Sentence-level vs. Session-level Context:} Due to the limitations of discrete feature extraction methods like TF-IDF, analysis often remains strictly at the isolated sentence or single-post level. Algorithms treat each tweet or comment as an independent event, completely neglecting the chronological and contextual dynamics of conversational threads. Cyberbullying is rarely a single isolated utterance; it is a sustained behavior. Machine learning models struggle to maintain state or track the escalating dynamics between aggressors, victims, and bystanders over a continuous session.
  \item \textbf{Textual vs. Multimodal Data:} Early machine learning models are fundamentally confined to textual processing. While some researchers attempt to concatenate numerical metadata (e.g., account age, follower count, posting frequency) with text features to create a pseudo-multimodal input, true integration of visual data (images, videos) is practically impossible due to the sheer mathematical complexity and the inability of traditional machine learning classifiers to process high-dimensional pixel data efficiently alongside text.
\end{itemize}

To address these semantic gaps and dimensionality curses, subsequent research begins incorporating distributed word representation techniques such as Word2Vec and GloVe. 
These continuous vector spaces not only capture latent semantic relationships between words, improving the detection of nuanced insults, but they also serve as a crucial historical bridge, ushering the field of cyberbullying detection into the era of deep learning.

\subsection{Deep Learning Methods on Cyberbullying Detection}
The transition from traditional machine learning to deep learning marks a paradigm shift from manual feature engineering to automated, hierarchical representation learning. Deep learning methods demonstrate significant advantages in cyberbullying detection because they can efficiently process large-scale datasets, automatically extract complex abstract features, and accurately classify both textual and visual instances \cite{hasan2023review}. 
Compared to traditional machine learning methods, Deep learning architectures accelerate the learning and feature selection processes, thereby greatly enhancing real-time detection capabilities on fast-paced social media platforms \cite{libina2023automatic}. 
In the critical tasks of identifying cyberbullying and spam comments, deep learning architectures consistently outperform traditional methods, highlighting their dominant superiority in this domain \cite{meenakshi2023deep}.

Architecturally, deep learning frameworks in this domain encompass supervised learning models—such as Deep Neural Networks (DNN), Multilayer Perceptrons (MLP), Convolutional Neural Networks (CNN), and the Recurrent Neural Network (RNN) family, which includes Bi-directional RNNs, Long Short-Term Memory (LSTM), and Gated Recurrent Units (GRU) \cite{abulwafa2022survey,cm2023deep,mosavi2019list,balakrishna2022comparative}. 
Extensive analyses of these deep learning methods reveal that supervised learning approaches consistently outperform unsupervised ones in social media cyberbullying detection, with the BiLSTM algorithm demonstrating particularly outstanding performance due to its bidirectional contextual awareness.
Crucially, the evolution of deep learning technologies catalyzes profound breakthroughs across the three core dimensions of cyberbullying detection:
\begin{itemize}
  \item \textbf{From Explicit to Implicit Detection:} 
  The introduction of sequential models like LSTMs equips algorithms with the fundamental ability to capture word order and deep semantic correlations. 
  Combined with attention mechanisms, models can transcend mere keyword matching to recognize semantic contradictions. 
  This contextual understanding enables deep learning models to identify implicit cyberbullying.
  For example, by capturing contextual dependencies, a BiLSTM with an attention mechanism can detect inherent sarcasm or mockery, drastically reducing the false-negative rates associated with covert microaggressions that previously evade lexicon-based machine learning filters.
  \item \textbf{From Sentence-level to Session-level Context:} 
  Deep learning facilitates the leap from isolated single-sentence analysis to session-level conversational modeling. 
  Researchers utilize deep neural networks to capture the temporal dynamics of multi-turn dialogues, proving that integrating contextual history is essential for identifying complex harassment behaviors where isolated comments only reveal their malicious intent when viewed through the lens of an ongoing session.
  \item \textbf{From Textual to Multimodal Data:} 
  Deep neural networks revolutionize the transition from text-centric analysis to comprehensive multimodal detection. While early research focuses strictly on plain text, modern architectures enable the integration of heterogeneous signals—including images, video, and psychological traits—to capture harm across diverse social media formats. 
  For instance, researchers successfully fuse NLP models with psychological features \cite{wang2025parental} and image-based sentiment \cite{kumari2025identifying} to uncover complex abuse that eludes text-only systems. 
  Specialized frameworks like CB-YOLO and hybrid BiLSTM-ResNet architectures further demonstrate the efficacy of cross-modal fusion in identifying derogatory memes and visual sarcasm \cite{dong2025follow, gupta2025novel}. 
  This progression currently extends into the video domain, where synchronizing audio-visual signals with real-time metadata remains an open challenge.
\end{itemize}

While deep learning architectures establish new state-of-the-art benchmarks, they still require extensive domain-specific fine-tuning and massive amounts of labeled data. 
Furthermore, their ability to perform deep, common-sense reasoning regarding cultural nuances remains constrained, setting the stage for the emergence of Large Language Models.

\subsection{LLMs for Cyberbullying Detection}

In recent years, cyberbullying detection enters a new phase driven by modern generative and instruction-tuned Large Language Models (LLMs). Unlike previous classifiers that often rely on task-specific feature engineering or large-scale supervised fine-tuning, generative LLMs such as GPT, Claude, Gemini, LLaMA, Mistral, Qwen, and DeepSeek can perform cyberbullying detection through natural-language instructions, in-context examples, and explanation-oriented prompting. This shift further improves the generalization ability of models in few-shot and zero-shot scenarios, while helping to address the limitations of previous approaches in implicit expression understanding, long-range contextual modeling, and multimodal content analysis~\cite{cirillo2025exploring,muminovic2025benchmarking}.

\begin{itemize}
  \item \textbf{From Explicit to Implicit Detection:}
  In real online environments, cyberbullying does not always appear in explicit forms such as direct insults, threats, or profanity. Instead, it is often conveyed through sarcasm, coded language, euphemisms, memes, indirect humiliation, and culturally specific online slang. These implicit forms of abuse may not contain obvious harmful keywords, making them difficult to identify through surface-level lexical matching alone~\cite{elsherief2021latent}. Modern generative LLMs are advantageous in this regard because they can use linguistic knowledge, pragmatic reasoning, and contextual understanding to identify subtle hostile intent and potential psychological harm. Through zero-shot, few-shot, and explanation-oriented prompting, LLMs can be instructed not only to determine whether a piece of content is harmful, but also to identify the abusive expression, the target of attack, the type of bullying, and the evidence supporting the decision~\cite{cirillo2025exploring}. Building on this direction, recent work further explores the incorporation of auxiliary signals, such as aggression-related cues, into prompts, allowing the model to use additional semantic information when detecting cyberbullying and thereby improving its generalization in complex scenarios~\cite{saeid-etal-2025-cyberbullying}.
 
  \item \textbf{From Sentence-level to Session-level Context:}
  Cyberbullying is often not fully captured by a single comment or sentence, but may emerge and escalate through continuous interactions. Modern LLMs, especially those with long-context capabilities, make it possible to analyze comments, conversation histories, and user interaction records as coherent discourse rather than as independent textual fragments. This modeling paradigm enables the model to track how harmful intent accumulates over time, distinguish one-time offensive speech from sustained bullying behavior, and identify the evolving roles of instigators, victims, and bystanders. Compared with sentence-level detection, session-level reasoning is more consistent with the social and psychological nature of cyberbullying, and it also helps avoid oversimplifying complex interactions into isolated binary classification tasks~\cite{verma-etal-2024-beyond}. In addition, LLMs can generate structured summaries of long conversations, highlight key turning points, and provide clearer contextual evidence for human moderators.

  \item \textbf{From Textual to Multimodal Data:}
  With the evolution of social media, cyberbullying increasingly occurs in multimodal online spaces, where harmful meaning may be distributed across text, images, emojis, memes, screenshots, short videos, livestream comments, and audio-visual interactions. In such scenarios, abusive intent may not be fully expressed in the text alone, but may arise from the semantic conflict between a visual object and its caption, a humiliating image edit, a mocking meme template, or hostile comments attached to otherwise neutral media. Multimodal LLMs and Vision-Language Models extend the scope of cyberbullying detection by jointly analyzing images, text, and other modal cues, enabling models to identify implicit harm from cross-modal semantic relationships. Recent studies begin to explore the use of LLM-generated multi-perspective explanations combined with smaller models for multimodal meme-based cyberbullying detection~\cite{joshi2026multimodal}. Such approaches show considerable potential for short-video platforms, image-centered social media, and livestreaming scenarios, although their practical effectiveness still depends on high-quality multimodal datasets, reliable cross-modal reasoning, and strict privacy-preserving mechanisms.
\end{itemize}

Despite these advances, the practical deployment of LLMs in cyberbullying detection remains challenging. Their predictions are sensitive to prompt design, model versions, decoding settings, and platform-specific language styles, which may lead to inconsistent results across systems. Existing datasets also suffer from class imbalance, cultural specificity, and limited coverage of low-resource languages and dialects, thereby weakening real-world generalization. Computational cost, inference latency, data privacy, hallucinated explanations, and accountability issues further constrain their use in content moderation~\cite{muminovic2025benchmarking,garciaMendez2025explainable}.

Future research should develop context-aware, explainable, and human-centered moderation frameworks that combine lightweight classifiers, LLM-based reasoning, and human review. Key directions include session- and user-level modeling, multimodal detection, multilingual and low-resource evaluation, adversarial robustness, privacy-preserving deployment, and evidence-grounded explanation generation. Rather than replacing human moderators, LLMs should serve as decision-support components for identifying high-risk content, organizing contextual evidence, estimating severity, and improving the transparency of intervention. Thus, their value lies not only in classification accuracy, but also in supporting more fine-grained, accountable, and socially responsible online safety governance.

\begin{figure}[t]
    \centering
    \includegraphics[width=\linewidth]{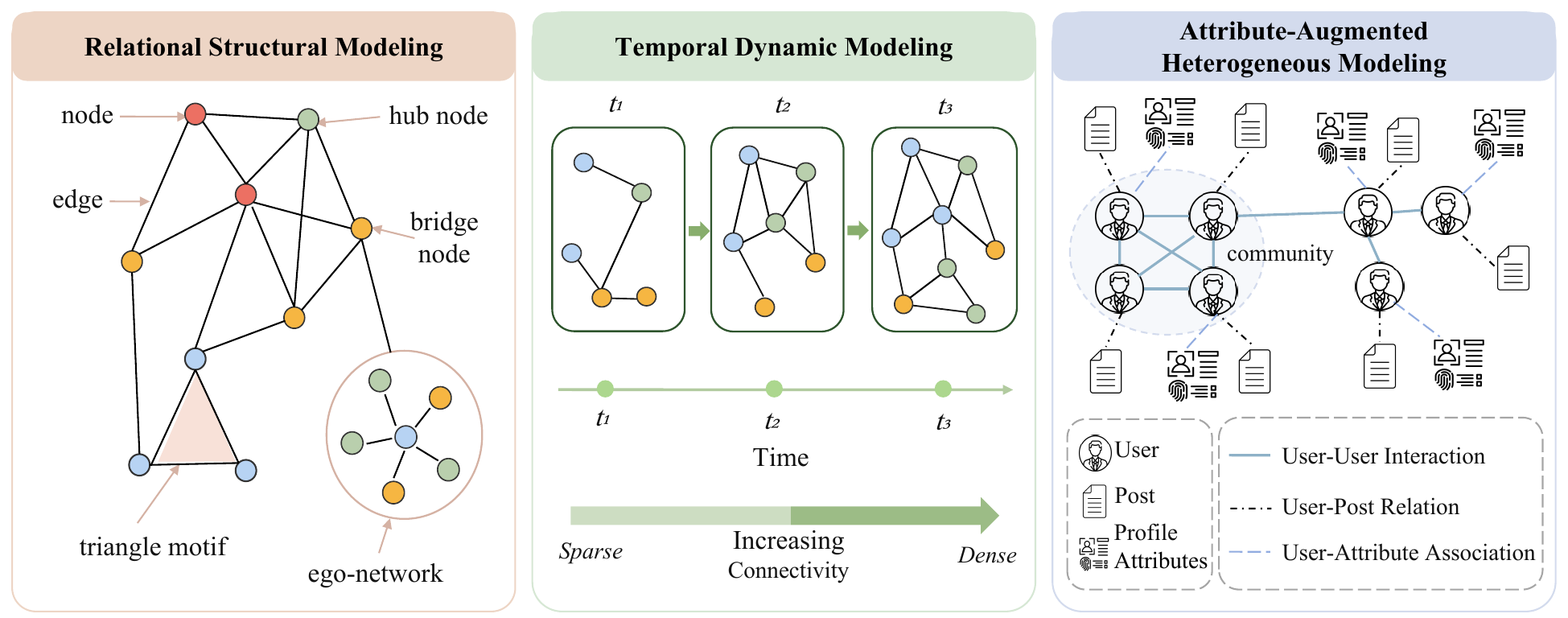}
    \caption{Taxonomy of social network modeling methods for cyberbullying governance.}
    \label{fig:social_network_modeling}
\end{figure}

\section{User and Behavior Modeling}\label{sec:behavior}

In cyberbullying governance, identifying harmful content from individual posts or comments alone is often insufficient to explain how harm emerges and escalates. 
Cyberbullying is better understood as a dynamic social process shaped by replies, reposts, coordinated participation, and collective amplification—raising questions of who initiates attacks, how they are reinforced through interaction, and how they propagate through social structures. 
We therefore organize this section into three closely related directions: social network modeling, role identification, and social bot detection.

\subsection{Social Network Modeling}\label{subsec:sn_modeling}

In cyberbullying governance, social network modeling focuses not on individual harmful messages themselves, but on how attacks are organized, reinforced, and propagated through relational structures and interaction chains. Because cyberbullying is often embedded in replies, mentions, reposts, and bystander participation, isolated text alone is often insufficient to capture its underlying formation mechanisms. 
Existing approaches can be broadly grouped into three directions, namely modeling based on relational generation mechanisms and structural statistics, modeling based on temporal evolution and dynamic graphs, and modeling based on attribute augmentation and heterogeneous graphs, as illustrated in Figure~\ref{fig:social_network_modeling}.

\begin{figure}[t]
    \centering
    \includegraphics[width=\linewidth]{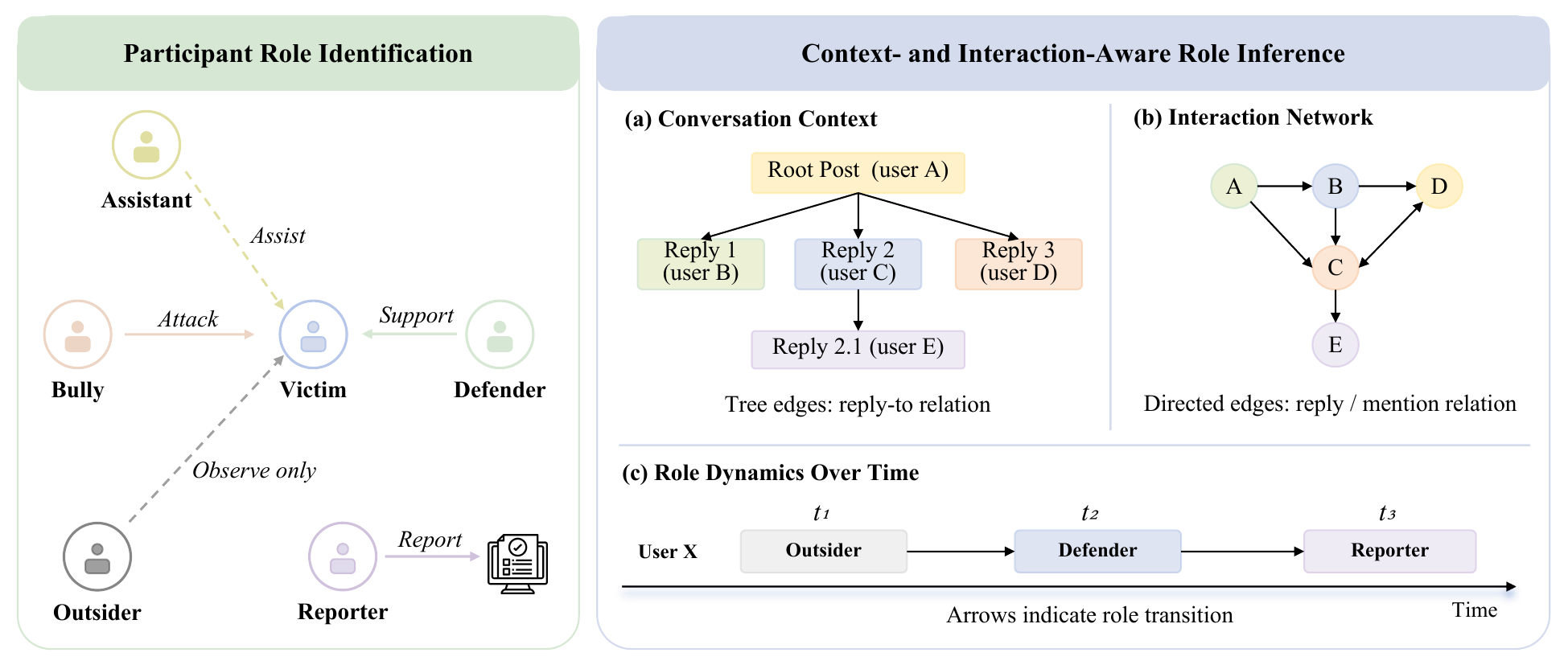}
    \caption{Taxonomy of role identification methods for cyberbullying governance.}
    \label{fig:role_identification}
\end{figure}

\textbf{Relational generation mechanisms and structural statistics.}
  This direction explains how networks form through local dependency patterns and structural statistics. 
  Representative foundations include exponential random graph models~\cite{holland1981exponential,frank1986markov,wasserman1996logit} and the Louvain community detection method~\cite{blondel2008fast}. 
  In cyberbullying settings, ~\citet{huang2014cyber} construct ego-networks around senders and receivers for joint structural and textual modeling, while~\citet{menini2019system} integrate message classification with social network analysis to track repeated attacks within specific communities. 
  This line of work is effective for identifying core aggressor clusters, high-risk communities, and anomalous relational patterns.

\textbf{Temporal evolution and dynamic graphs.} 
  This direction focuses on the escalation process of cyberbullying by incorporating edge activation time into network analysis~\cite{holme2012temporal}. 
  Dynamic graph representation learning methods such as JODIE~\cite{kumar2019predicting} and DyRep~\cite{trivedi2019dyrep} learn how node embeddings evolve over time. 
  In cyberbullying applications,~\citet{chelmis2019minority} formulate Instagram cyberbullying prediction as a multi-horizon intensity forecasting problem, and \citet{ge2021improving} propose a unified temporal graph that jointly models topical consistency across comments and temporal interaction dynamics within a single graph framework.
  Compared with static methods, this direction is better suited to identifying concentrated attacks, relay-like pile-ons, and the progression from emergence to escalation.

\textbf{Attribute augmentation and heterogeneous graphs.} 
  This direction jointly incorporates text, attributes, communities, and relational structure into a unified framework. 
  General technical foundations include Attributed Social Network Embedding~\cite{liao2018attributed}, R-GCN~\cite{schlichtkrull2018modeling}, and HAN~\cite{wang2019heterogeneous}. 
  In cyberbullying-related applications,~\citet{mishra2019abusive} use GCNs on author-post graphs for abusive language detection, \citet{kurrek2022enriching} show that community context improves abusive expression modeling, and \citet{chowdhury2019arhnet} propose ARHNet, which combines community interaction graphs with textual representations to detect Arabic religious hate speech.
  The advantage of this line of work is that it captures not only what is being said, but also who says it, which community the speaker belongs to, and how the speaker interacts with others. 
  It is therefore better suited to identifying pile-on behavior in which textual signals may appear normal while relational structure remains anomalous.

Overall, relational structural modeling is effective for identifying central actors and high-risk communities. 
Temporal dynamic modeling is better suited to characterizing escalation processes. 
Attribute-augmented heterogeneous modeling is better aligned with complex risk identification in real platform environments. 
For cyberbullying governance, the significance of social network modeling lies not merely in providing additional features, but in explaining why pile-ons emerge, how they escalate, and which structures are most worthy of intervention from the perspectives of relation formation, interaction dynamics, and multi-source joint modeling.

\subsection{Role Identification}\label{subsec:roles}





Role identification focuses not on whether a single message is harmful, but on who is performing which function in a cyberbullying incident—who is attacking, who is being targeted, who is reinforcing the attack, who is offering support, and who is merely observing. 
This shift moves the analytical focus from harmful text to participant behavior, and thus provides a more actionable basis for intervention. 
In practice, role identification can help platforms distinguish users who require immediate protection from those who should be prioritized for enforcement, while also identifying bystanders who may be guided toward constructive responses. 
Existing studies have generally evolved through three directions, as illustrated in Figure~\ref{fig:role_identification}.

\textbf{Coarse-grained role categorization.} 
 Early work usually adopts a bully--victim--bystander framework and formulates role identification as a multiclass text classification task. 
 This line of research shows that cyberbullying detection should not be limited to recognizing abusive content alone, but should also account for the functional roles expressed in social media text \cite{van2018automatic}. 
 Subsequent studies further demonstrate the feasibility of automatically identifying victims, bullies, and bystanders from textual traces, making it clear that role recognition is a distinct task rather than a simple extension of binary cyberbullying detection \cite{jacobs2022automatic}. 
 Related work also shows that incorporating participant-role information can improve cyberbullying identification itself, suggesting that role labels are not merely descriptive, but can also strengthen event-level understanding \cite{rathnayake2020enhancing}.

\textbf{Fine-grained sub-role modeling.} 
  Subsequent work further refines the bystander category into more specific roles such as defender, assistant, outsider, and reporter—reflecting the fact that bystanders are not behaviorally homogeneous: some amplify harm, some mitigate it, and others remain passive~\cite{polanco2021bystander}. 
  More recent computational studies likewise explore multi-role settings involving harassers, victims, assistants, defenders, and other bystander types, and show that finer-grained labeling is more suitable for supporting targeted moderation and intervention, even though it also introduces greater class imbalance and semantic ambiguity \cite{sandoval2024identifying,zhong2025beyond}. 
  
\textbf{Contextual and interaction-aware modeling.} 
  Recent work recognizes that participant roles often cannot be inferred reliably from a single comment in isolation, as the same utterance may function differently depending on its conversational context, reply relation, or position in the interaction sequence. 
  For this reason, recent work increasingly incorporates session context, reply structure, and user interaction patterns into role inference.
  \citet{kao2019understanding} propose a social role detection framework for Instagram and Ask.fm that explicitly models victim, bully, and victim-supporter roles through social and linguistic signals, while \citet{verma2024beyond} argue that participant roles should be modeled jointly with conversational structure rather than as flat labels detached from context. 
  This direction indicates that role identification is gradually evolving from sentence-level classification toward context- and relation-driven modeling.

Overall, role identification has progressed from coarse-grained participant classification to fine-grained, context-aware behavioral modeling, providing a more operational understanding of who drives harm, who suffers it, and who amplifies or mitigates it. 
At the same time, several challenges remain: publicly available role-annotated datasets are still limited, class imbalance is common, role boundaries are often ambiguous, and participant roles may shift as an incident unfolds. 
Future work should therefore move toward more dynamic, interaction-aware, and governance-oriented modeling frameworks.

\begin{table*}[t]
\centering
\footnotesize
\caption{Summary of representative social bot detection methods for cyberbullying governance.}
\label{tab:social_bot_detection}
\renewcommand{\arraystretch}{1.18}
\begin{tabular*}{\textwidth}{@{\extracolsep{\fill}}llllcccl@{}}
\toprule
\multirow{2}{*}{\textbf{Method}} &
\multirow{2}{*}{\textbf{Year}} &
\multirow{2}{*}{\textbf{Category}} &
\multirow{2}{*}{\textbf{Dataset}} &
\multicolumn{3}{c}{\textbf{Modalities}} &
\multirow{2}{*}{\textbf{Core Technique}} \\
\cmidrule(lr){5-7}
& & & & \textbf{Meta.} & \textbf{Text} & \textbf{Graph} & \\
\midrule
Kudugunta et al.~\cite{kudugunta2018deep} 
& 2018 & Text-based & Cresci-2017 
& \checkmark & \checkmark &  & Deep neural network \\

Wei et al.~\cite{wei2019twitter} 
& 2019 & Text-based & Cresci-2017 
&  & \checkmark &  & BiLSTM with word embeddings \\

DeepBot~\cite{luo2020deepbot} 
& 2020 & Text-based & Cresci-2017 
&  & \checkmark &  & Deep neural network \\

Dukić et al.~\cite{dukic2020you} 
& 2020 & Text-based & PAN 2019 English 
&  & \checkmark &  & BERT encoder \\

Heidari et al.~\cite{heidari2020using} 
& 2020 & Text-based & Cresci-2017 
&  & \checkmark &  & BERT-based sentiment features \\

Graph-Hist~\cite{magelinski2020graph} 
& 2020 & Graph-based & Conversational graphs 
&  &  & \checkmark & Latent feature histogram classification \\

BotRGCN~\cite{feng2021botrgcn} 
& 2021 & Graph-based & TwiBot-20 
& \checkmark & \checkmark & \checkmark & Relational graph convolution \\

RGT~\cite{feng2022heterogeneity} 
& 2022 & Graph-based & TwiBot-20 
& \checkmark & \checkmark & \checkmark & Relational graph transformer \\

Dehghan et al.~\cite{dehghan2023detecting} 
& 2023 & Graph-based & Italian Election 
&  &  & \checkmark & Node and structural embeddings \\

BIC~\cite{lei2023bic} 
& 2023 & Multimodal & Cresci-15, TwiBot-20 
& \checkmark & \checkmark & \checkmark & Text-graph interaction \\

BotMoE~\cite{liu2023botmoe} 
& 2023 & Multimodal & Cresci-15,TwiBot-20/22 
& \checkmark & \checkmark & \checkmark & Community-aware MoE \\

DGT~\cite{he2024dynamicity} 
& 2024 & Graph-based & TwiBot-20/22 
& \checkmark & \checkmark & \checkmark & Dynamic graph transformer \\

LGB~\cite{zhou2025lgb} 
& 2025 & Multimodal & TwiBot-20/22  
& \checkmark & \checkmark & \checkmark & LM and GNN fusion \\
\bottomrule
\end{tabular*}
\end{table*}

\subsection{Social Bots Detection}\label{subsec:bots}

In cyberbullying governance, social bot detection focuses on identifying anomalous accounts that amplify attacks, manipulate interaction dynamics, and manufacture false consensus through automated or semi-automated behavior. 
Such accounts shape pile-on structures through abnormal replies, reposts, and coordinated interactions, thereby intensifying harm and accelerating escalation. 
The significance of this task therefore lies not merely in identifying suspicious accounts per se, but in revealing the actors that actively drive the spread and amplification of online attacks. 
Along this line, existing research has gradually evolved from text-based methods to graph-based methods and, more recently, multimodal fusion methods, with representative works summarized in Table~\ref{tab:social_bot_detection}.

\textbf{Text-based methods.} 
Early approaches rely on user descriptions, posts, and linguistic style to identify social bots, using deep neural networks, recurrent neural networks, and word embeddings to model textual expression~\cite{kudugunta2018deep,wei2019twitter,luo2020deepbot}. 
Later studies introduce pretrained language models to strengthen the modeling of user semantics and sentiment \cite{dukic2020you,heidari2020using}. 
In the context of cyberbullying, such methods are particularly useful when complete social graphs are unavailable, since they allow platforms to perform early screening of suspicious accounts based on aggressive language, discourse style, and content-level anomalies alone. 
Their limitation, however, is that advanced bots may evade detection by reusing human-written posts or diluting malicious content, making text-only approaches unreliable against highly disguised accounts.



\textbf{Graph-based methods.} 
Rather than focusing on what an account says, this direction detects bots through social relations and coordinated behavior. 
Early work extracts topological signals through node and structural embeddings~\cite{dehghan2023detecting,magelinski2020graph}, while graph neural networks are now the dominant paradigm—BotRGCN uses heterogeneous relation graphs and relational graph convolution to aggregate information across relation types~\cite{feng2021botrgcn}, and relational graph transformers further model influence differences across heterogeneous relations~\cite{feng2022heterogeneity}.  
Recent work also extends this line to dynamic settings, using dynamic graph transformers to capture the temporal evolution of bot behavior \cite{he2024dynamicity}. 
For cyberbullying governance, graph-based methods are especially important as pile-on events are structured through abnormal reply chains, concentrated reposting, and coordinated amplification that text-based approaches fail to capture.
Compared with text-based methods, graph-based approaches are therefore better suited to identifying botnets or troll-like coordination, and to revealing which accounts drive pile-ons, bridge diffusion, or collectively amplify attacks.

\begin{figure}[t]
    \centering
    \includegraphics[width=\linewidth]{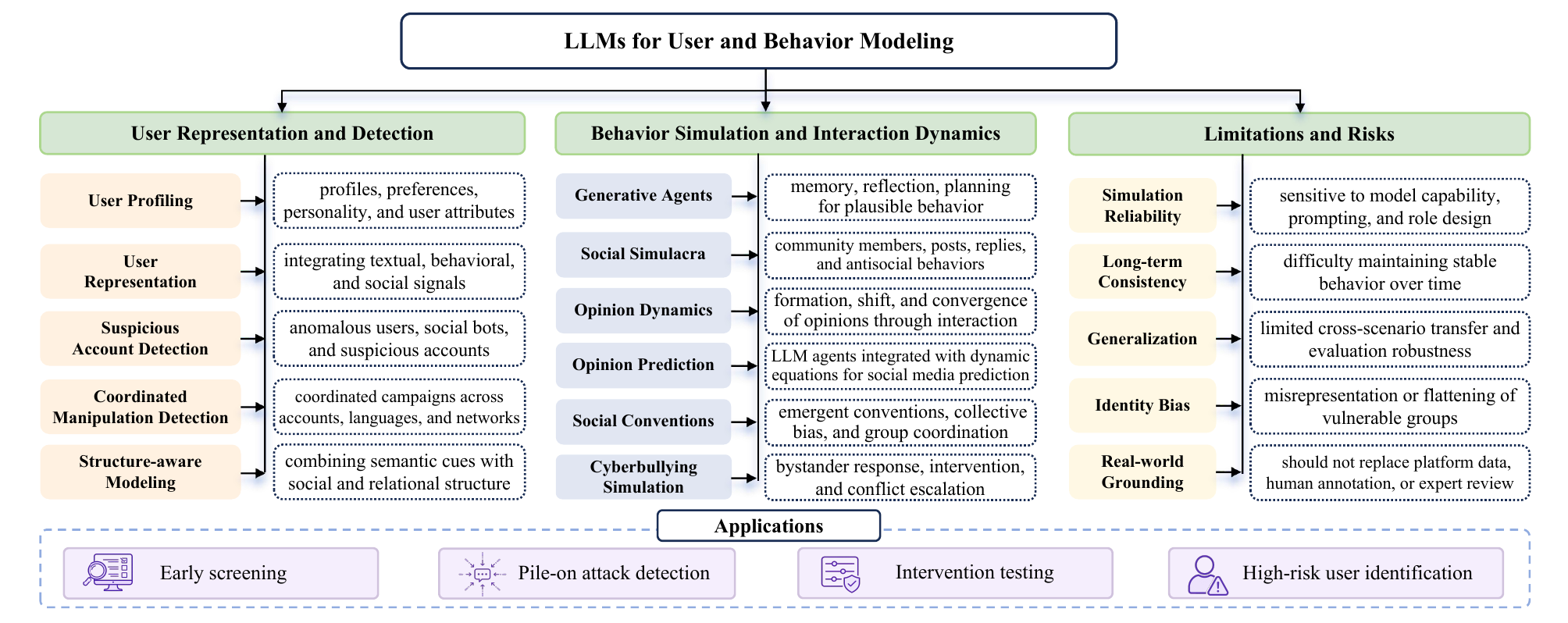}
    \caption{LLMs for User and Behavior Modeling.}
    \label{fig:llm_user_behavior}
\end{figure}

\textbf{Multimodal fusion methods.} 
This direction combines metadata, text, and graph information to overcome the limitations of single-modality detection. 
The key intuition is that bot accounts often expose inconsistency or coordinated abnormality across modalities, and joint modeling can therefore characterize disguise strategies more comprehensively. 
BIC improves performance through text-graph interaction and semantic consistency modeling, highlighting the importance of coupling relational structure with semantic representation \cite{lei2023bic}. 
BotMoE further introduces a community-aware mixture-of-experts mechanism to jointly model metadata, text, and network structure, improving robustness to community variation and highly disguised bots \cite{liu2023botmoe}. 
LGB extends this line by integrating language models with graph neural networks, further strengthening the complementarity between semantic and structural signals \cite{zhou2025lgb}.

Overall, social bot detection has evolved from text-based paradigms to graph-based modeling and multimodal fusion. 
In cyberbullying governance, its role should not be reduced to account-level classification; more importantly, it helps identify which accounts are amplifying attacks and manufacturing false consensus. 
Key challenges remain in three areas: the adversarial nature of bot evolution, the difficulty of generalizing across time, platforms, and datasets~\cite{feng2021twibot,yang2020scalable,yang2022botometer}, and the growing prevalence of human-bot hybrids and LLM-enabled disguise strategies that demand more robust multimodal and group-level coordination analysis. 
The most valuable future direction therefore lies in integrating anomalous-account detection into pile-on identification, diffusion warning, and intervention decision-making.

\subsection{LLMs for User and Behavior Modeling}\label{subsec:Modeling}
Large language models (LLMs) offer new opportunities for user and behavior modeling in cyberbullying governance. Because cyberbullying often involves implicit hostility, multi-turn interactions, coordinated participation, and shifting group attitudes, governance systems must capture not only abusive expressions but also user intent, behavioral patterns, and interaction dynamics. LLMs enrich user representations from profiles, posts, metadata, and social contexts; they identify suspicious or coordinated accounts via semantic reasoning, and simulate how users respond, align with others, or escalate conflicts during online interactions.
At the same time, deploying LLMs in cyberbullying governance introduces risks, including unreliable behavioral simulation, biased representation of vulnerable groups, and limited capacity for explicit relational structure modeling. 
Figure~\ref{fig:llm_user_behavior} summarizes this emerging direction from user representation and detection, behavior simulation and interaction dynamics, and the limitations and risks of LLM-based behavior modeling.

\textbf{LLMs for user representation and detection.}
General user modeling studies summarize the potential of LLMs for user profiling, preference understanding, personality modeling, and behavioral analysis, centered on learning richer user representations by integrating textual signals, user‑generated content, and graph structure \cite{tan2023user}.
For large‑scale social network settings, LLM‑based user modeling further combines language representations with social network data to capture both semantic characteristics and neighborhood‑level behavioral patterns \cite{jiang2025social}.
In anomalous account detection, recent methods use LLMs to jointly model user posts, metadata, and structural cues for social bot detection \cite{feng2024does}.
Related work extends LLM‑based detection to influence campaigns by jointly leveraging user metadata, network structure, and multilingual content to identify coordinated manipulation \cite{luceri2024leveraging}.
Further methods integrate LLM‑derived semantic representations with graph Transformers to jointly model account‑level content features and relational structures \cite{li2025botlgt}.
For cyberbullying governance, these methods enable platforms to move beyond atomic content analysis, identifying suspicious users, social bots, and coordinated manipulation groups from profiles, behavioral traces, and social relations — thereby supporting early malicious‑account screening, pile‑on attack detection, and high‑risk user group identification.

\textbf{LLM-based agents for behavior simulation and interaction dynamics.}
Foundational agent studies employ memory, reflection, and planning mechanisms to enable LLM‑driven agents to generate plausible individual behaviors and emergent social interactions \cite{park2023generative}.
In social computing systems, LLMs are also used to generate community members, posts, replies, and antisocial behaviors, making it possible to simulate how online communities evolve under different rules and interventions \cite{park2022social}.
For opinion and collective behavior modeling, LLM‑agent populations simulate opinion dynamics in social networks, capturing how individual opinions form, shift, and converge through interaction \cite{chuang2024simulating}.
Other studies combine LLM‑based agents with dynamic equations to predict opinion evolution on social media \cite{yao2025social}.
LLM populations are further used to examine how social conventions, collective bias, and group coordination emerge from agent interactions \cite{ashery2025emergent}.
More directly relevant to cyberbullying, LLM‑powered multi‑agent social media simulations analyze why young bystanders hesitate to speak up in cyberbullying scenarios and how simulated practice encourages more constructive bystander intervention \cite{yang2026attention}.
For cyberbullying governance, LLM‑based agents can model not only individual user responses but also multi‑user interactions, group attitude shifts, bystander participation, and conflict escalation — providing a controllable environment for intervention testing, platform rule design, bystander education, and early‑warning mechanism optimization.

\textbf{Limitations and risks of LLM-based behavior modeling.}
LLM‑based behavior modeling still faces reliability, fidelity, and fairness risks.
First, LLM‑based human simulation is sensitive to model capability, role specification, prompt design, and simulation framework, making it difficult to fully reproduce real human behavior \cite{wang2025limits}.
Second, LLM‑agent simulation remains limited in environmental perception, action generation, long‑term consistency, evaluation criteria, and cross‑scenario generalization ~\cite{gao2024large, mou2026individual}.
Moreover, replacing real users with LLMs in social behavior analysis may misrepresent or flatten identity groups, weakening the understanding of vulnerable populations, victims, and culturally specific contexts \cite{wang2025large}.
From the perspective of computational social science, LLMs are better viewed as tools for assisted modeling and hypothesis generation, not as substitutes for real platform data, human annotation, or rigorous social‑scientific validation \cite{ziems2024can}.
Thus, for cyberbullying governance, LLM‑based behavior modeling should be used together with real interaction data, graph‑structured modeling, human expert assessment, and fairness auditing, so as to avoid governance errors caused by simulation distortion, group bias, or insufficient relational modeling.

Overall, LLMs extend user and behavior modeling from traditional feature representation and task‑specific detection toward semantic reasoning, multi‑source user understanding, and simulation‑based interaction analysis.
User representation and detection methods help identify suspicious accounts, social bots, and coordinated manipulation, while LLM‑based agents provide a controllable environment for analyzing user interaction, opinion evolution, and potential conflict escalation.
However, LLM‑based behavior modeling remains constrained by limited simulation reliability, weak relational structure modeling, biased representation of identity groups, and insufficient grounding in real platform data.
Future research should integrate LLMs with graph‑structured modeling, real behavioral data, human expert validation, and fairness‑aware evaluation to support more robust, interpretable, and governance‑oriented analysis of users and behaviors in cyberbullying.

\begin{figure*}[!t]
\centering
\includegraphics[width=\textwidth]{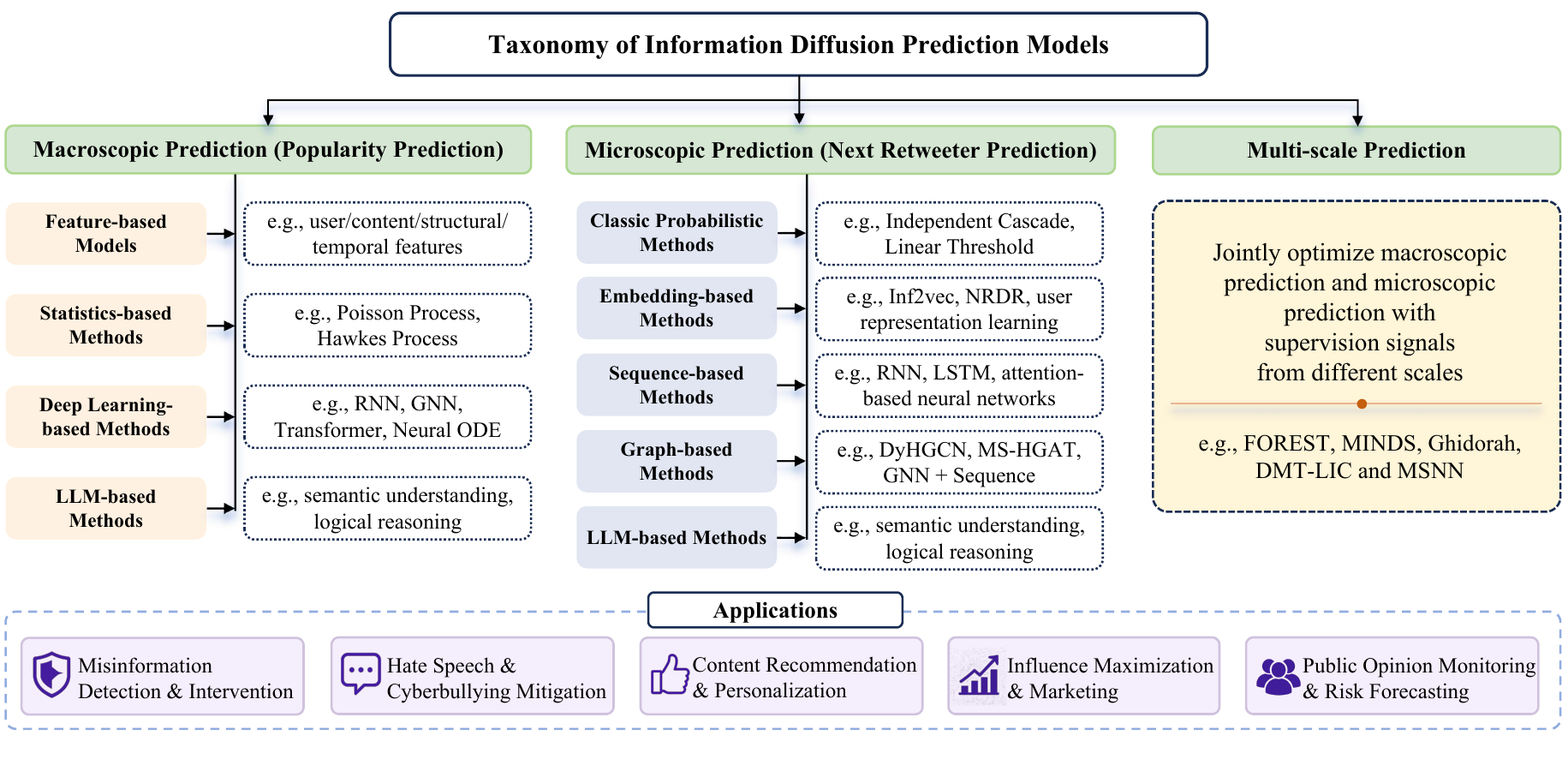}
\caption{Taxonomy of information diffusion prediction methods categorized by prediction granularity.}
\label{fig:idp_taxonomy}
\end{figure*}

\section{Diffusion Dynamics and Early Warning}\label{sec:propagation}

\subsection{Information Diffusion Prediction}\label{subsec:cascade}



Online social platforms such as Twitter, Sina Weibo, Facebook, etc., have brought unprecedented convenience for the production and dissemination of information. Understanding the underlying patterns of information diffusion on these platforms has significant societal and economic implications: it is pivotal not only for enabling platform moderators and relevant authorities to intervene early to curb the spread of misinformation~\cite{budak2011limiting, sun2023fighting}, hate speech~\cite{masud2021hate, goel2023hatemongers}, and cyberbullying content~\cite{mladenovic2021cyber}, but also for recommending high-quality content to relevant users and improving user engagement on the platform~\cite{alsoubai2024systemization}. To this end, researchers have devoted significant efforts to information diffusion prediction (IDP) task, which aims to predict future diffusion dynamics of information from observed social interaction records. 
From the perspective of prediction granularity, current IDP methods are typically categorized into two types \cite{DMT-LIC, li2023grass, wang2023multiscale, jiao2024enhancing}: \textit{macroscopic prediction} (a.k.a. popularity prediction)\cite{chen2019information, xu2021casflow, bao2024popularity}, which aims to estimate the number of retweets a post will receive after a certain time period, and \textit{microscopic prediction} \cite{wang2022cascade,sun2022ms,qiao2023rotdiff}, which aims to identify the next potential retweeter. These two types of tasks are traditionally studied in isolation~\cite{xu2021casflow, sun2022ms, feng2022h, bao2024popularity}, whereas recent works propose to employ a unified framework to jointly optimize both for \textit{multi-scale prediction}~\cite{yang2021full, jiao2024enhancing}. A taxonomy of representative IDP methods is illustrated in Figure~\ref{fig:idp_taxonomy}. In what follows, we review these three lines of research in turn.


\textbf{Macroscopic prediction} aims to predict the future popularity of an information cascade based on its early-stage diffusion patterns \cite{cheng2024information}. Existing methods can be broadly summarized into the following three categories \cite{cheng2024information, bao2024popularity}. (1) Feature-based methods. Early studies focus on extracting various hand-crafted features, e.g., user-related, content-related, structural, and temporal features, for popularity prediction \cite{cheng2014can, shulman2016predictability, chen2020event}. However, they heavily depend on feature quality and often lack generalizability. (2) Statistics-based methods. These approaches regard information diffusion as the arrival process of retweet events and characterize cascade growth by modeling the retweet intensity with Poisson \cite{shen2014modeling} or Hawkes \cite{cao2017deephawkes, rizoiu2017expecting} point processes, offering interpretability but suffering from limited predictive power. 
(3) Deep learning-based methods. As the current mainstream paradigm, these approaches focus on learning expressive cascade representations, capturing the complex temporal and structural dependencies within each cascade through architectures such as RNNs~\cite{li2017deepcas, chen2019information}, GNNs~\cite{cao2020popularity, bao2024popularity}, Transformers~\cite{chen2022and, wang2025casformer}, Neural ODEs~\cite{cheng2024information}, and related variants.


\textbf{Microscopic prediction} focuses on modeling individual users' responses to specific information, aiming to identify the user most likely to repost at the next step by ranking repost probabilities over users who have not yet participated in the cascade. Early studies directly estimate diffusion probabilities using predefined diffusion models such as Independent Cascade (IC) and Linear Threshold (LT)~\cite{saito2008prediction, goyal2010learning}. Later, embedding-based methods \cite{bourigault2016representation, feng2018inf2vec, wang2019information} encode diffusion dependencies of cascades into user embeddings and compute diffusion probabilities through well-designed functions. Both classes of methods are constrained by restrictive assumptions and tedious feature extractions. Given the sequential nature of diffusion cascades, subsequent studies propose to exploit recurrent neural networks (RNNs) to predict the next potential retweeter \cite{wang2017topological, islam2018deepdiffuse, wang2018sequential}. Despite notable progress, such approaches focus solely on diffusion dependencies among users, while ignoring the influence of underlying social relations. To address this limitation, recent approaches typically combine graph neural networks (GNNs) with sequence models to jointly capture the social network structure and the temporal ordering within cascades~\cite{sun2022ms, qiao2023rotdiff, cheng2023enhancing, wang2024information, zhong2024information}.





Motivated by the complementary and mutually reinforcing nature of the two tasks, recent efforts begin to explore unified frameworks for multi-scale prediction that jointly model both macroscopic and microscopic diffusion dynamics~\cite{yang2021full, jiao2024enhancing, zhu2025ghidorah}.
Despite the progress outlined above, several challenges remain open. 
First, most existing models, especially deep learning-based ones, operate as black boxes, offering limited interpretability regarding why a cascade spreads in a particular way. 
Second, current methods primarily rely on cascade-related data to capture diffusion patterns, while overlooking a wide range of contextual influencing factors such as discourse context, emotional resonance, and value identity. Third, most studies consider cascades in isolation, whereas in practice multiple cascades often co-exist and interact, competing for user attention or reinforcing each other, and modeling such competitive and cooperative dynamics remains largely underexplored. Finally, current research is mostly evaluated on general-purpose datasets (e.g., Weibo, Twitter, APS), while dedicated benchmarks for cyberbullying propagation are still lacking, constraining the development of prediction models tailored to such scenarios.


\subsection{Opinion Dynamics and Polarization Analysis}\label{subsec:dynamics}



When a controversial event emerges on social media, public discussion rarely remains static. 
The focal topics shift as the event develops, individual and collective opinions evolve under social influence, and the affective distance between opposing camps may progressively widen. 
Research on this temporal evolution can be organized around three interconnected sub-problems: \textit{topic evolution detection and tracking}, \textit{opinion dynamics modeling}, and \textit{polarization analysis and prediction}. 
From the perspective of cyberbullying governance, these three lines of inquiry are jointly indispensable: understanding the dynamic mechanisms driving opinion polarization provides the analytical foundation for predicting when a controversy is escalating from ordinary disagreement into collective hostility, enabling timely intervention.


\textbf{Topic evolution detection and tracking.} 
Tracking how discussion topics shift across the lifecycle of a social media event has traditionally relied on probabilistic topic modeling. 
Latent Dirichlet Allocation (LDA) \cite{blei2003latent} and its temporal extensions—the Dynamic Topic Model (DTM) \cite{blei2006dynamic} and Topics Over Time (TOT) \cite{wang2006topics}—capture phase-level topic changes by decomposing text corpora at successive time windows. 
For short social media texts, the Biterm Topic Model (BTM) \cite{cheng2014btm} addresses data sparsity by modeling word co-occurrence across documents. 
More recently, BERTopic \cite{grootendorst2022bertopic} has become the dominant neural approach, encoding documents with sentence embeddings and supporting dynamic topic tracking across user-defined time bins. 
Extensions such as BERTrend \cite{largeron2024bertrend} further leverage LLMs to automatically summarize topic developments between consecutive timestamps. 
In the cyberbullying context, these methods can reveal how discussion migrates from the triggering incident toward the personal details of the target or questions of platform responsibility—marking the semantic trajectory of an escalating pile-on.


\textbf{Opinion dynamics modeling.} 
The agent-based modeling (ABM) tradition provides the theoretical backbone, ranging from discrete models such as the Voter model \cite{holley1975ergodic} and the Sznajd model \cite{sznajd2000opinion}, to continuous models including DeGroot averaging \cite{degroot1974reaching}, the Hegselmann--Krause bounded-confidence model \cite{hegselmann2002opinion}, and the Friedkin--Johnsen model \cite{friedkin1990social}, which introduces stubbornness to generate persistent disagreement consistent with empirical observations. 
On the data-driven side, epidemic-inspired compartmental models—particularly SEIR variants—have been widely adopted to simulate public opinion propagation by analogy with disease diffusion \cite{zhang2025seir}. 
The GAN-SEIR model \cite{ganseir2025} further combines a Generative Adversarial Network with the SEIR framework, addressing independence assumptions and parameter-fixity limitations of classical formulations. 



\textbf{Polarization analysis and prediction.} 
A key distinction separates \textit{issue polarization}—the bimodal distribution of positions on a specific question—from \textit{affective polarization}—the intensification of hostility toward out-groups \cite{iyengar2015fear}. 
Echo chambers and filter bubbles constitute the structural preconditions of polarization in many theoretical accounts. 
A systematic analysis of 112 studies finds that the scholarly disagreement on echo chambers stems from divergent operationalizations: data-driven computational studies grounded in homophily tend to confirm the echo chamber hypothesis, while content-exposure studies tend to reject it \cite{hartmann2025echo}. 
On the measurement side, network-structural approaches based on homophily, modularity, and random walks represent the dominant family of polarization metrics \cite{interian2023network}, complemented by content-based methods applying sentiment analysis and stance detection. 
Empirically, in-group social media interactions exhibit positive affect while out-group replies are characterized by negativity and toxicity, with this gradient extending continuously as a function of network distance \cite{chen2024affective}. At the causal level, a preregistered field experiment on X during the 2024 US presidential campaign demonstrated that algorithmically reducing partisan animosity exposure shifted out-party hostility by over two points on a 100-point scale \cite{jia2025reranking}.


Despite substantial progress, several critical gaps remain. First, topic evolution and opinion dynamics have been studied largely in isolation, with no unified framework jointly modeling how content shifts and stance changes interact. Second, hybrid models such as GAN-SEIR and FDE-LLM are promising but have been validated primarily in single-platform settings, limiting generalizability. Third, LLM agents exhibit an inherent bias toward consensus with accurate information \cite{chuang2024simulating}, limiting their utility for modeling opinion radicalization. Finally, the computational escalation pathway from polarization to targeted harassment remains systematically underexplored, leaving a crucial gap in early-warning and intervention system design.



\begin{figure*}[!t]
\centering
\includegraphics[width=\textwidth]{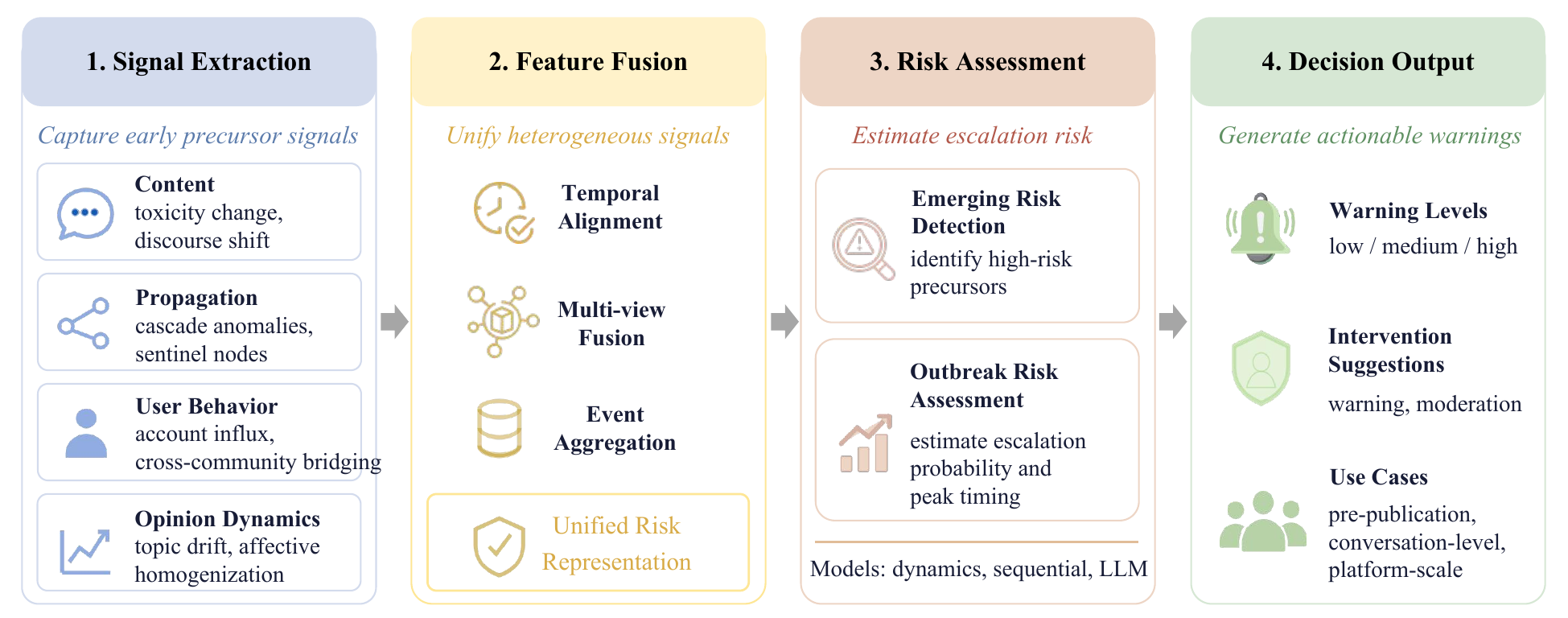}
\caption{Four-layer framework of an early warning system (EWS) for cyberbullying. }
\label{fig:EWS}
\end{figure*}

\subsection{Early Warning Systems}\label{subsec:early_warning}


Early warning systems (EWS) for cyberbullying occupy the decision-output layer of the predictive pipeline, synthesizing the propagation-scale models of Section~\ref{subsec:cascade} and the opinion-dynamics frameworks of Section~\ref{subsec:dynamics} into actionable, tiered risk assessments for pre-emptive intervention. 
As illustrated in Figure~\ref{fig:EWS}, this pipeline comprises four stages: signal extraction, feature fusion, risk assessment, and decision output. 
The core task decomposes into two logically nested sub-problems: emerging risk detection, which identifies high-risk precursors from the continuous content stream, and outbreak risk assessment, which quantifies escalation probability and peak timing for an already-identified event.
This two-stage structure distinguishes EWS from conventional single-pass toxicity classifiers and motivates the multi-signal, temporally grounded modeling approaches reviewed below.


\textbf{Signal extraction.} 
Detecting nascent cyberbullying requires integrating precursor signals across four analytical layers. 
At the content layer, the per-post rate of change in toxicity scores—rather than their absolute level—is a more reliable precursor than static offensiveness~\cite{zhang2018conversations}, and discourse shifts such as increased second-person targeting consistently precede conversational derailment~\cite{chang2019trouble}.
In the propagation layer, cascade structural anomalies that include anomalous branching factors and the participation signal of sentinel-nodes coordinate amplification independently of content toxicity~\cite{hosseinmardi2016prediction,li2023grass}. 
The user behavior layer adds account-influx anomalies and cross-community bridging as documented precursors to large-scale campaigns~\cite{im2024stoking}, while the opinion dynamics layer contributes spiral-of-silence suppression and affective homogenization-indicators that observable discourse systematically underestimates actual mobilization~\cite{muller2023hashtag}. 

\textbf{Feature fusion and risk assessment.} 
Translating these signals into risk estimates has followed three modeling strategies. Propagation-dynamics approaches adapt compartmental epidemiological models to toxicity spread, enabling counterfactual scenario analysis for intervention timing~\cite{milli2025evaluating}. Deep sequential models offer greater flexibility: Chang and Danescu-Niculescu-Mizil's CRAFT architecture~\cite{chang2019trouble} reframed derailment as an emergent conversational property, learning turn-level dynamic representations that outperform static baselines; subsequent work extended this line with dynamic BERT training~\cite{kementchedjhieva2021dynamic} and LLM-based zero-shot prediction~\cite{olpadkar2025can}. 
The third strategy concerns tiered alert design: translating continuous risk scores into graded decisions requires calibrating the trade-off between false positives and false negatives under asymmetric intervention costs. 
Twitter's pre-publication nudge exemplifies this tension, reducing harmful content through a 9\% deletion and 22\% revision rate while raising concerns about the suppression of legitimate critical speech~\cite{katsaros2022reconsidering}.



\textbf{Decision output and tooling.} 
The methods above have been operationalized across three categories. 
Pre-publication detection embeds classifiers at the point of content creation, as seen in Twitter and Tinder nudge systems~\cite{katsaros2022reconsidering}. 
Conversation-level prediction issues continuously updated assessments as threads develop, with evidence that 64\% of toxic exchanges become detectable within 24 hours of the first derailment signal~\cite{imran2025toxicity}. 
Platform-scale scoring services such as Jigsaw's Perspective API provide production-grade coverage but remain limited to single-feature threshold architectures~\cite{schluger2022proactive}. 
Together, these approaches reflect an inherent tension between lead-time and accuracy, as earlier intervention points operate with less observed evidence and necessarily accept higher uncertainty.


The EWS field nevertheless faces five structural challenges. The dominant paradigm evaluates models at the post level rather than across the full incident trajectory, creating a mismatch with the operational unit of harm. The four signal classes lack a unified temporal alignment framework. 
The boundary between cyberbullying and legitimate adversarial discourse remains ill-defined, making precision-recall calibration inherently contested. 
Burst-type incidents impose a cold-start constraint that is fundamentally at odds with the observation windows required for stable signal  estimation~\cite{lopez2021early}. 
Finally, existing corpora are heavily concentrated on English single-platform data, leaving cross-lingual and cross-platform generalization largely unaddressed. Future work should prioritize LLM-based fine-grained intent modeling with systematic bias evaluation, causal inference frameworks for measuring intervention efficacy, and cross-platform architectures capable of tracking harassment campaigns across their full multi-platform lifecycle.


\subsection{LLMs for Diffusion and Early Warning}\label{subsec:LLM_for_diffusion_warning}

Traditional methods for modeling diffusion, opinion evolution, and escalation risk mainly rely on network structures, temporal patterns, and behavioral statistics. While effective in capturing large-scale propagation and interaction patterns, they often overlook the semantic and pragmatic content of online interactions. This limitation is particularly critical in cyberbullying scenarios, where subtle cues such as sarcasm, implicit hostility, emotional resonance, and multimodal expressions can directly shape message propagation, opinion radicalization, and harm escalation. Recent studies therefore introduce LLMs into diffusion prediction, social simulation, and early warning systems, leveraging their semantic understanding and reasoning capabilities to enrich conventional models with linguistic and contextual signals.

\textbf{LLMs for diffusion prediction.} Recent studies~\cite{chen2025large, zheng2025autocas, xu2025forecasting, kayal2025large, shang2026make, liu2026llm} begin to explore LLMs as semantic and reasoning engines for information diffusion prediction. Unlike conventional models that mainly rely on cascade structures or early engagement signals, LLMs can provide auxiliary cues such as user influence, latent social roles, topical virality, audience resonance, and multimodal semantics. Representative works include TLGM~\cite{chen2025large}, which combines heterogeneous graph representations with LLMs for topic-level short-video peak prediction; LSID~\cite{liu2026llm}, which introduces LLM-derived Semantic IDs for semantic-aware user modeling; and AutoCas~\cite{zheng2025autocas}, which reformulates cascade popularity prediction as an autoregressive next-token prediction task and repurposes LLMs as cascade predictors. MILD~\cite{shang2026make} further leverages LLM reasoning to infer explainable who-influences-whom relations, while BuzzProphet~\cite{xu2025forecasting} and recent video popularity studies~\cite{kayal2025large} use LLM-generated rationales or VLM--LLM pipelines to enrich popularity forecasting. 
These works indicate that LLMs are particularly useful for injecting semantic priors and causal rationales into diffusion models. However, challenges remain in aligning semantic, structural, and behavioral signals, as well as in ensuring numerical stability, controlling hallucinations, improving computational efficiency, and constructing multimodal benchmarks specifically designed for cyberbullying diffusion scenarios.

\textbf{LLMs for opinion simulation.} Recent LLM-based agent studies~\cite{mou2024unveiling, chuang2024simulating, yang2024oasis, mi2026mf, liu2025popsim, huang2026policysim, zhang2026intervensim} explore how LLMs can simulate opinion dynamics, polarization, and collective attitude shifts in social media environments. HiSim~\cite{mou2024unveiling} adopts a hybrid design where influential core users are modeled by LLM agents and massive ordinary users by ABMs, enabling scalable simulation of opinion-leader behavior and social movement responses. OASIS~\cite{yang2024oasis} builds a generalizable social-media sandbox with dynamic networks, recommendation mechanisms, rich action spaces, and million-scale agents to reproduce information spreading, group polarization, and herd effects. Recent works further emphasize realism and intervention: MF-LLM~\cite{mi2026mf} uses mean-field agent--population interactions for scalable alignment with real decision dynamics; PopSim~\cite{liu2025popsim} simulates UGC propagation via social mean fields for popularity prediction; PolicySim~\cite{huang2026policysim} trains social agents with SFT/DPO and optimizes recommendation or exposure-control policies through sandbox feedback; and IntervenSim~\cite{zhang2026intervensim} models source-side interventions and crowd deliberation to capture consensus--polarization transitions. 
These studies show that LLMs can serve as opinion agents, counterfactual simulators, and policy-test environments, but still face challenges such as prompt-dependent personas, heuristic memory, high cost, weak calibration, and consensus bias, which may limit their ability to capture contentious, non-consensual, and adversarial opinion dynamics.

\textbf{LLMs for Early Warning.} In proactive moderation, LLMs facilitate a critical shift from static toxicity classification to dynamic trajectory forecasting. Research indicates that well-designed zero-shot and few-shot prompting strategies, when combined with chain-of-thought reasoning, allow LLMs to achieve competitive forecasting performance on derailment benchmarks without requiring task-specific fine-tuning \cite{olpadkar2025can}. Going a step further, generative approaches treat the future of a conversation as a distribution to be sampled \cite{zhang2025forecasting}. By generating multiple plausible continuations of a thread via a fine-tuned LLM and utilizing majority voting, such methods have yielded absolute accuracy improvements of 4--7\% on the CGA-Wiki dataset and 18--20\% on BNC. At the utterance level, providing fine-grained contextual information—such as target community identities—further enhances zero-shot detection performance \cite{roy2023probing}. Despite these improvements, detecting implicit hostility remains an open problem. Current models frequently misclassify ambiguous terms or identity-laden vocabulary in benign contexts as harmful, producing poorly calibrated confidence scores for nuanced interactions \cite{zhang2024don}.

\textbf{Challenges.} The inherent consensus bias of standard LLMs severely limits their capacity to model adversarial social dynamics. Aligned rigorously to prioritize safety and factual rationality, these models inherently struggle to spontaneously simulate the polarization and radicalization processes characteristic of real-world cyberbullying. Recent studies observe that simulated agent populations consistently converge toward cooperative, objective discourse rather than exhibiting human-like entrenchment or affective contagion \cite{chuang2024simulating}. Addressing this gap remains an open problem, requiring novel alignment techniques or prompting frameworks to effectively model irrational escalation.

\begin{figure*}[!t]
\centering
\includegraphics[width=\textwidth]{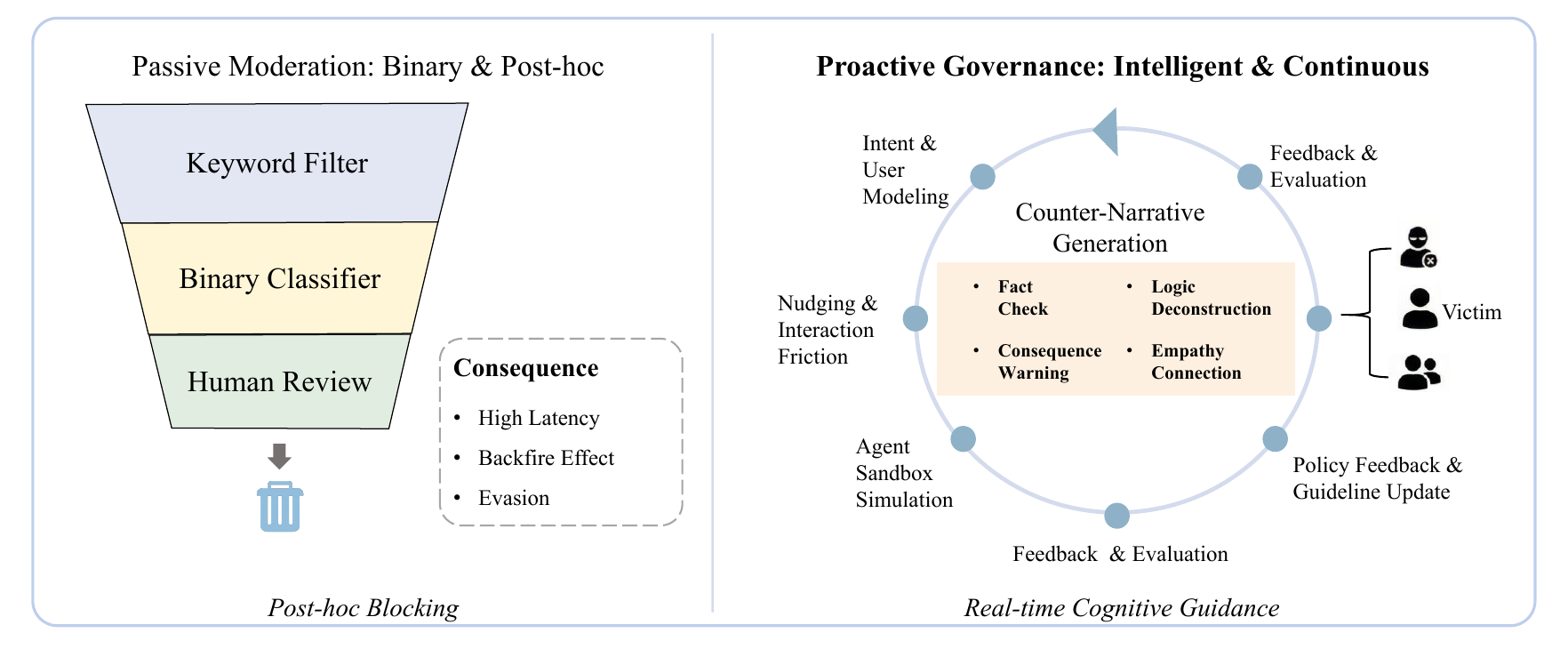}
\caption{Web Content Governance Models Comparison: Passive Funnel vs. Intelligent Loop}
\label{fig:CFBully}
\end{figure*}

\section{Intervention and Proactive Governance}
\label{sec:intervention}

\subsection{Paradigm Shift: From Passive Moderation to Proactive Governance}
\label{subsec:paradigm_shift}

Currently, mainstream social media platforms primarily rely on ex-post moderation and punitive measures, such as content deletion, account suspension, or shadowbanning, to combat cyberbullying. While these binary blocking mechanisms are essential for filtering explicit hate speech and meeting compliance requirements, their limitations become evident as cyberbullying grows more collective and covert \cite{gillespie2020content}. The primary issue with passive moderation is latency. Because intervention inherently lags behind the abuse, malicious content often spreads rapidly before batch processing or manual review completes, causing immediate psychological harm. Moreover, binary moderation is often too rigid for nuanced disputes, leaving platforms caught between inaction and over-censorship~\cite{gorwa2020algorithmic}. Relying solely on punitive measures can also backfire, inciting coordinated retaliation or prompting users to adopt adversarial strategies, such as using homophones, emojis, or image-based text to bypass algorithmic detection, which further complicates governance.

To address these limitations, recent research suggests shifting from passive defense to proactive algorithmic governance, as illustrated in Figure~\ref{fig:CFBully}. 
This transition reflects an evolution in governance goals, moving beyond simple content removal toward conflict de-escalation and community restoration. Simultaneously, advances in Large Language Models (LLMs) now enable real-time, semantic, and interactive interventions. Proactive governance reshapes moderation across three areas. Temporally, it allows platforms to intervene before conflicts escalate, addressing the latency of ex-post methods. Methodologically, it shifts the focus from binary punishment to timely nudges and intelligent responses. Finally, the overall objective expands from deleting violative content to guiding online behavior and establishing community norms.

Within this proactive framework, automated Counter-Narratives (CN) serve as a core intervention mechanism. Unlike traditional warnings, counter-narratives provide responses grounded in factual clarification, logical deconstruction, and empathy. They help repair the community ecosystem in several ways. For impulsive users, presenting objective facts or potential consequences immediately after an offensive post creates a cognitive buffer, reducing the likelihood of further emotion-driven escalation. For victims, public counter-narratives offer immediate social support by refuting harmful statements, which is often more effective than the psychological isolation caused by silent content deletion. Furthermore, visible counter-narratives signal to bystanders that malicious behavior is unacceptable, mitigating the broken windows effect caused by accumulated toxic comments. By moving from ex-post punishment to real-time cognitive guidance, counter-narratives encourage the community's capacity for self-correction and establish a foundation for modern automated governance.

\subsection{Automated Counter-Narrative Generation}
\label{subsec:counter_narrative}

Understanding proactive algorithmic governance requires clarifying the role of Counter-Narratives (CN) as an independent intervention mechanism. Unlike standard content moderation or system prompts, counter-narratives avoid rigid violation warnings and direct adversarial arguments with aggressors. Instead, they aim to correct misinformation, de-escalate conflicts, support victims, and positively influence bystanders, thereby integrating content generation, stance expression, and interactive governance. Traditional generation baselines, which often rely on static templates or unconstrained Large Language Models (LLMs), struggle with highly toxic inputs. To address these limitations, recent work has proposed generation architectures tailored for severe cyberbullying contexts~\cite{chung2019conan, safdari2026parscn}. These robust intervention systems typically integrate Retrieval-Augmented Generation (RAG), Direct Preference Optimization (DPO), and multi-agent pre-intervention simulation.

Designing these systems begins with structuring the intervention space. Literature typically categorizes counter-narrative strategies into four primary orientations: factual correction, logical deconstruction, consequence warning, and empathetic connection. These orientations are not mutually exclusive. In practice, a single counter-narrative often combines multiple elements, such as pairing factual clarification with emotional support. Modern system designs operationalize this by using a front-end intent recognition module to estimate the probability distribution of these target strategies, which then constrains the back-end generation model to synthesize text that reflects a specific composite strategy.

\begin{figure*}[!t]
\centering
\includegraphics[width=\textwidth]{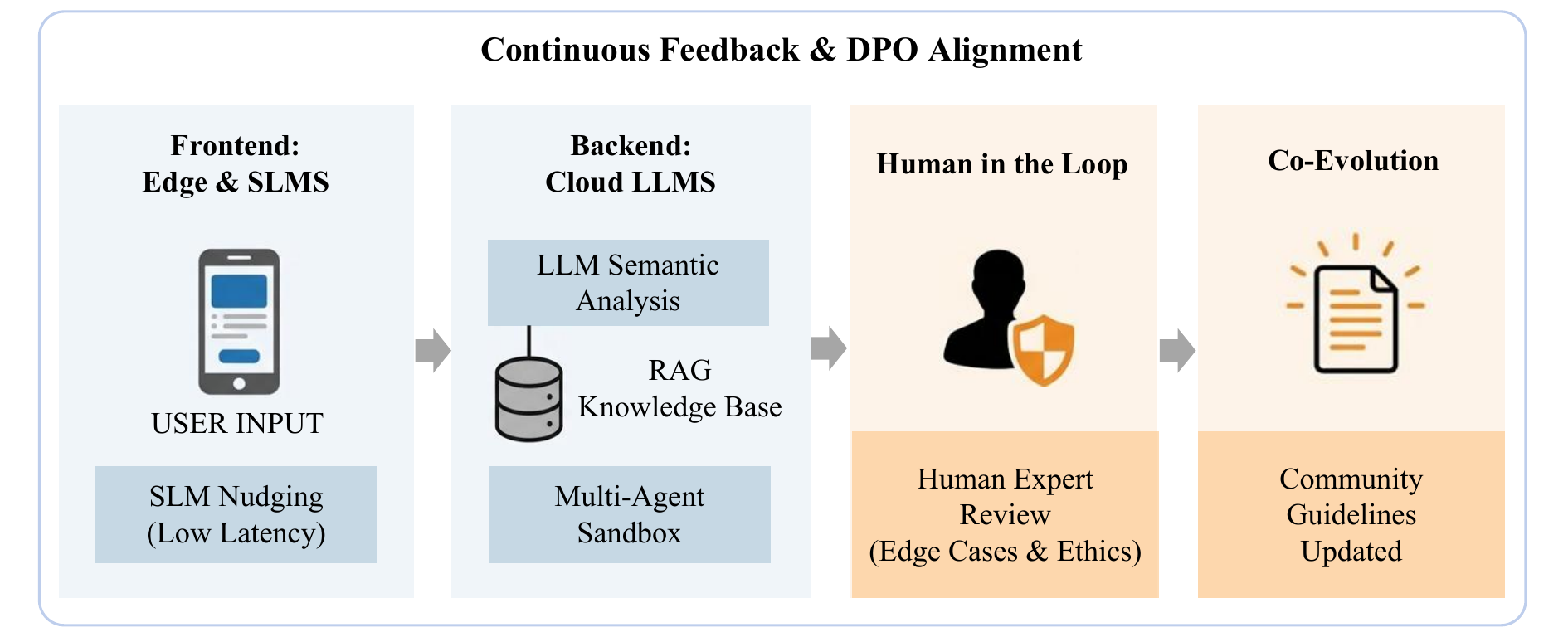}
\caption{A Co-Evolutionary Hybrid AI System with Integrated RAG and Continuous Feedback}
\label{fig:HybridAI}
\end{figure*}

\subsection{Content Moderation and Policy Optimization}
\label{subsec:moderation_policy}

The efficacy of proactive governance depends on how effectively automated generation technologies integrate with existing moderation systems. Recent academic discussions focus on a collaborative human-AI approach that combines behavioral science, tiered computing architectures, and human oversight~\cite{fanton2021human, gajewska2026improving}. This framework transitions away from fragmented moderation by establishing a continuous pipeline: frontend prevention, mid-tier routing, backend intervention, human review, and feedback iteration. This systematic approach aims to balance computational efficiency with ethical fairness, as depicted in Figure~\ref{fig:HybridAI}.

At the frontend, conflict prevention focuses on influencing user interactions before abuse occurs, shifting away from reactive moderation. Research highlights the utility of algorithmic nudging for this purpose. If a system detects emerging aggressive speech, it can introduce cognitive friction, such as prompting the user to consider milder phrasing, which helps interrupt emotional escalation. For bystanders, nudges can encourage them to support victims or endorse counter-narratives, activating the community's capacity for self-repair. Ethically, these mechanisms must be strictly limited to mitigating impulsive, emotion-driven harm. They should not attempt to discipline users' core viewpoints, ensuring the platform does not engage in ideological policing.

To manage industrial-scale data throughput while balancing intervention depth, latency, and computational costs, systems typically employ a tiered computing architecture based on task risk. For high-frequency, low-risk tasks, small language models deployed at the frontend perform millisecond-level lightweight screening, risk scoring, and templated nudging without disrupting the user experience. When incidents involve complex contexts or higher risks, the system routes the tasks to cloud-based large language models. These backend models execute deep semantic analysis, generate personalized counter-narratives as described in Section \ref{subsec:counter_narrative}, and run multi-agent sandbox simulations. Finally, if a situation presents extreme semantic ambiguity or severe threats of real-world harm, the system triggers a circuit breaker, immediately escalating the ultra-high-risk case to a manual review channel. This stratification makes proactive governance practical for large-scale deployment.

Even with efficient automated screening, models risk biased misjudgments when handling edge cases that involve cultural taboos, sarcasm, or complex interpersonal dynamics. Within this pipeline, human experts provide the core legitimacy for platform governance. While automated systems handle the initial processing of massive datasets, human moderators are responsible for nuanced value judgments, contextual interpretation, and ultimate accountability. This human-in-the-loop framework ensures transparency and maintains continuous ethical oversight over machine interventions.

Continuous feedback iteration enables the platform ecosystem to adapt over time. By collecting data from human-machine collaboration, particularly the corrections human experts apply to model outputs, backend algorithms can dynamically learn the evolving characteristics of cyberbullying, including novel coded language and coordinated attack strategies. This learning loop improves model accuracy and facilitates a co-evolution of technology and institutional policy. The risk trends identified by the algorithms provide empirical support for updating community guidelines, refining appeal processes, and adjusting macro-governance metrics. Consequently, the platform shifts from executing static rules to continuously optimizing its moderation policies.


\subsection{LLMs for Intervention and Governance}
\label{subsec:llm_intervention_governance}

In cyberbullying intervention, the value of LLMs does not lie only in generating fluent responses. Their main contribution is the ability to connect contextual interpretation, response generation, policy reasoning, and human oversight within the same governance pipeline. Unlike conventional moderation models, which usually classify content after harm has already occurred, LLM-based systems can examine conversational context, infer communicative intent, and support interventions before a conflict further escalates. This property is especially relevant to cyberbullying, where harmful interaction is often implicit, relational, and highly dependent on local context.

LLMs can serve as the generative core of automated counter-narratives. However, when an intervention involves factual correction or consequence warning, relying only on the parametric knowledge of an LLM may produce unsupported claims and weaken the credibility of the response. Recent systems therefore incorporate RAG as a grounding mechanism~\cite{lewis2020retrieval}. Platform guidelines, fact-checking resources, and community norms can be retrieved as external evidence to improve factual consistency and make generated responses more verifiable. Nevertheless, retrieval does not remove all uncertainty. Its effectiveness depends on the quality, recency, and cultural coverage of the retrieved sources. In cases involving sarcasm, value conflict, or relational humiliation, external evidence can support factual grounding, but it cannot fully resolve normative ambiguity.

Another important direction is the alignment of LLM outputs toward constructive and de-escalatory intervention. In adversarial discussions, an unconstrained model may imitate the hostility of the input or generate a response that intensifies the conflict. Recent work therefore moves beyond surface-level toxicity filtering and explores preference-based alignment methods, such as DPO~\cite{rafailov2023direct, yong2024preference}. By contrasting constructive counter-narratives with toxic or provocative alternatives, preference tuning can shift the output distribution toward empathy, clarification, and conflict reduction. The central difficulty is that constructiveness is not a universal standard. It varies across platforms, languages, and cultural communities. As a result, the quality of alignment depends on how preference data are collected, whose judgments are represented, and whether minority linguistic and cultural contexts are adequately included.

LLMs also make it possible to evaluate interventions through multi-agent sandboxing before deployment. Instead of directly posting a generated response in a live discussion, a system can simulate possible reactions from aggressors, victims, bystanders, and moderators~\cite{park2023generative}. Such simulation allows a platform to compare alternative intervention strategies, estimate the risk of backlash, and select a more conciliatory response when necessary. This changes intervention design from open-loop generation to anticipatory evaluation. However, simulated agents remain limited approximations of real users. Their behavior is sensitive to prompt design, and their ability to generalize across platforms, languages, and social contexts remains uncertain.

At the system level, LLMs can strengthen human--AI collaborative moderation by supporting semantic routing, risk stratification, and continuous policy feedback. On large platforms, lightweight models can handle low-risk nudging at the frontend, while LLMs can analyze high-context cases, summarize relevant evidence for human reviewers, and map incidents to applicable community guidelines~\cite{franco2025integrating, chen2025comprehensive}. Human experts remain indispensable for ambiguous, culturally sensitive, or high-stakes cases, but LLMs can reduce review burden by organizing context and suggesting possible intervention paths. Reviewer corrections and appeal outcomes can then be used to update model alignment and platform policy, forming a feedback loop between technical systems and institutional governance.

\begin{table*}[t]
\centering
\caption{Dataset Statistics}
\label{tab:dataset statics}

\tiny
\setlength{\tabcolsep}{2.5pt}
\renewcommand{\arraystretch}{1.1}

\begin{tabularx}{\textwidth}{l c c c c c c c c c}
\toprule
Dataset & Platform & Size & Proportion & Year & Collection Method & Annotation Method & Session-based & Language & Multimodal \\
\midrule

\citet{d1} & Myspace & 4813 & 0.21 & 2011 & Crawled & Research Assistant & $\checkmark$ & English &  \\
\citet{d2} & Youtube & 3468 & 0.14 & 2014 & Crawled & Research Assistant &  & English &  \\
\citet{d3} & Ins & 2215 & 0.29 & 2015 & Crawled & Crowdsourced & $\checkmark$ & English &  \\
\citet{d4} & Vine & 970 & 0.31 & 2015 & Crawled & Crowdsourced & $\checkmark$ & English &  \\
\citet{d5} & Twitter & 62587 & 0.05 & 2019 & API & Research Assistant &  & English &  \\
\citet{d6} & Twitter & 48000 & 0.83 & 2020 & API & - &  & English &  \\
\citet{d7} & Twitter & 5792 & 0.48 & 2021 & API & Research Assistant &  & Hinglish &  \\
\citet{d8} & Twitter & 12428 & 0.3 & 2022 & Crawled & Research Assistant &  & Urdu &  \\
\citet{d9} & Ins & 46898 & 0.26 & 2022 & API & Research Assistant &  & Arabic &  \\
\citet{d10} & AskFm & 10000 & 0.21 & 2022 & Crawled & Research Assistant &  & English &  \\
BullySentEmo\cite{d11} & Twitter & 6084 & 0.49 & 2022 & API & Research Assistant &  & Hinglish & $\checkmark$ \\
MultiBully\cite{d12} & Twitter, Reddit & 5854 & 0.55 & 2022 & Crawled & Research Assistant &  & Hinglish & $\checkmark$ \\
CYBY23\cite{d13} & Twitter & 112 & 0.66 & 2023 & API & Crowdsourced & $\checkmark$ & English &  \\
AlgD\cite{d14} & Facebook, YouTube, Twitter & 14150 & 0.31 & 2023 & API & Research Assistant &  & Multilingual &  \\
\citet{d15} & Twitter & 11041 & 0.16 & 2024 & API & Research Assistant &  & Polish &  \\
MC-Hinglish1.0\cite{d16} & Twitter & 8400 & 0.82 & 2024 & API & Research Assistant &  & Hinglish &  \\
BullyExplain\cite{d17} & Twitter & 6084 & 0.5 & 2024 & API & Research Assistant &  & Hinglish & $\checkmark$ \\
CHCIN\cite{d18} & Douyin, Weibo, Xiaohongshu, Bilibili & 220676 & 0.19 & 2025 & Crawled & Research Assistant &  & Chinese &  \\
SCCD\cite{d19} & Weibo & 677 & 0.52 & 2025 & Crawled & Research Assistant & $\checkmark$ & Chinese &  \\
CYBY24\cite{d20} & Twitter & 13309 & 0.21 & 2026 & API & Research Assistant & $\checkmark$ & English &  \\
pccd\cite{d21} & Twitter & 1668 & 0.26 & 2026 & API & Research Assistant & $\checkmark$ & Cantonese-English &  \\

\bottomrule
\end{tabularx}
\end{table*}

Despite these opportunities, LLM-based intervention still faces substantial risks. Hallucinated evidence can damage trust in counter-narratives. Preference alignment may encode dominant cultural assumptions and marginalize minority norms. Autonomous intervention may also generate unexpected social consequences if it is deployed without sufficient human oversight. In addition, the same generative capacity that supports constructive intervention can be misused to produce evasive abuse, coordinated harassment, or persuasive manipulation. Future research should therefore prioritize verifiable generation, culturally diverse alignment data, auditable human-in-the-loop workflows, and realistic multi-agent evaluation environments. In this sense, the role of LLMs is not simply to generate better text, but to support adaptive, accountable, and context-aware governance for cyberbullying intervention.

\section{Cyberbullying Datasets}\label{sec:datasets}
\subsection{Dataset Overview}\label{subsec:benchmarks}

To provide a comprehensive overview of existing cyberbullying datasets, we collected publicly available cyberbullying datasets from prior studies and open repositories. To ensure quality and comparability, datasets were included only if they (1) explicitly target cyberbullying detection, (2) provide annotated labels, and (3) are accessible for research purposes. In total, 21 datasets were selected, covering multiple social media platforms such as Twitter, Instagram, and Reddit. These datasets exhibit substantial diversity in terms of scale, annotation strategies, and data organization. 
Given the breadth of available resources, we focus on a representative subset rather than aiming for exhaustive coverage.

Table \ref{tab:dataset statics} summarizes the key characteristics of the collected datasets, including their sources, sizes, and annotation methodologies. In addition to these basic attributes, we further categorize the datasets based on their structural organization into two types: non-session-based data (e.g., individual posts or comments) and session-based data.

\begin{figure}[t]
  \includegraphics[width=0.9\columnwidth]{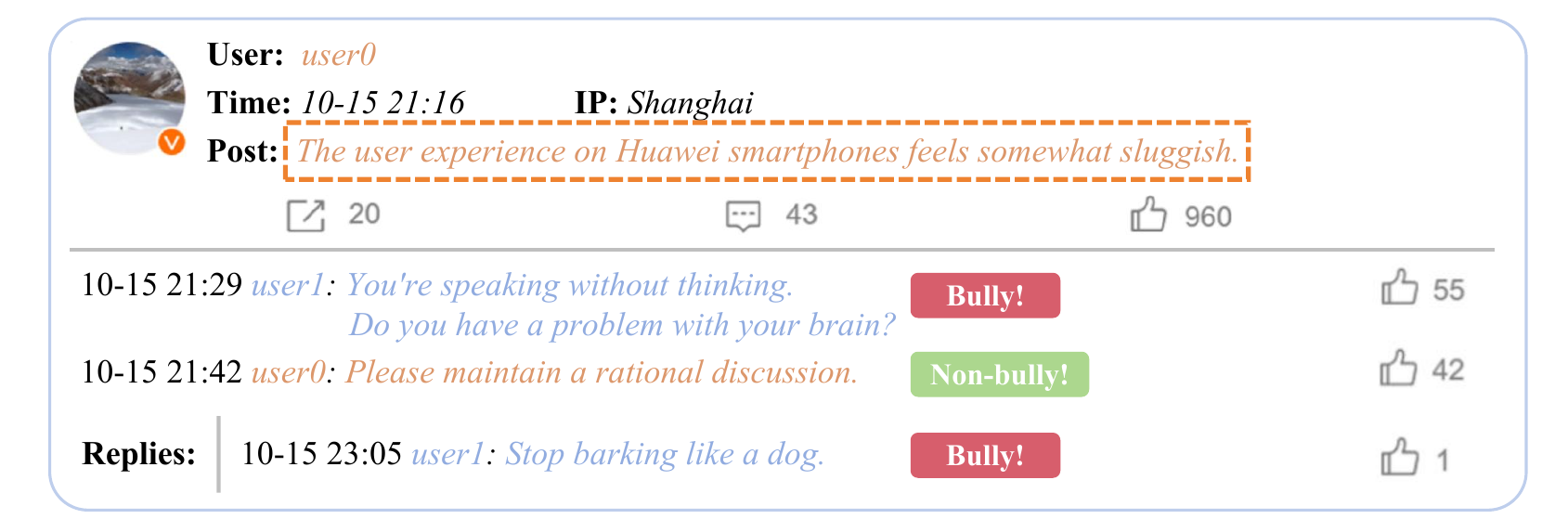}
  \caption{An illustration of a cyberbullying session from Weibo.}
  \label{fig:session-example}
\end{figure}

Non-session-based data typically consist of isolated textual units such as posts, tweets, or standalone comments. In contrast, session-based data is organized around a central post along with its associated responses, forming interaction threads among multiple users. Such structures often include richer, multi-granular information, such as user relationships, reposting behavior, user profiles, and engagement signals (e.g., likes or replies). To illustrate the structure of session-based data, Figure \ref{fig:session-example} presents an example of a session instance, highlighting the organization of posts, replies, and associated metadata.

\subsection{Cross-Dataset Analysis}\label{subsec:metrics}
Based on the datasets summarized in Table \ref{tab:dataset statics}, we conduct a systematic analysis to better understand their overall characteristics. Although these datasets vary considerably across multiple aspects, they also exhibit several common patterns. In the following, we analyze them from multiple perspectives, including data distribution, data sources and coverage, modality, as well as structural characteristics.

The majority of datasets exhibit significant class imbalance, where cyberbullying instances typically account for only a small proportion of the data, often below 30\%. In some extreme cases, the imbalance is particularly severe; for example, in the dataset introduced by \citet{d5}, the proportion of bullying samples is as low as 0.05. Such distributions are generally consistent with real-world scenarios, reflecting the inherent rarity of cyberbullying behavior compared to normal interactions. Nevertheless, a few datasets present a relatively high proportion of cyberbullying instances, sometimes exceeding 50\%\cite{d6,d12,d13,d16,d19}. This is often the result of deliberate data construction strategies, such as targeted sampling or filtering from high-risk or sensitive topics, rather than naturally occurring distributions.

The collected datasets exhibit substantial diversity in their sources, reflecting the heterogeneous nature of social media environments. They are drawn from a variety of platforms, including Twitter, Facebook, and Weibo, most of which are globally dominant. 
Notably, 13 out of the 21 datasets are sourced from Twitter, indicating a strong reliance on this platform and highlighting its central role in cyberbullying research. 
However, this diversity is somewhat superficial, as most datasets remain confined to a single platform, with relatively limited cross-platform resources.

From a linguistic perspective, existing datasets are predominantly English-centric. 
Among the 21 datasets examined, 16 are constructed in English, accounting for the vast majority of available resources, while datasets in other languages, such as Chinese and Arabic, remain relatively scarce. 
More recently, increasing attention has been paid to code-mixed data, including Indian English and Cantonese–English mixtures, reflecting the complexity of real-world online communication. Nevertheless, such resources are still limited in scale.

In terms of modality, current datasets are largely unimodal, with a primary focus on textual content. Only a small number of datasets incorporate multimodal information, such as images and emojis. Although recent efforts have begun to explore multimodal settings—for example, the MultiBully dataset\cite{d12}, which integrates both textual and visual data—multimodal resources remain at a relatively early stage of development.

A closer examination of data structure reveals that most datasets are non-session-based, typically consisting of isolated posts, tweets, or comments. Approximately 66\% of the collected datasets fall into this category. While such data are convenient for modeling, they fail to capture the continuity of user interactions. In contrast, session-based datasets, which model conversational or interactional dynamics, are less common. Due to the complexity and cost of constructing and annotating such data, they are generally smaller in scale, often comprising only hundreds to a few thousand instances. Despite these challenges, there has been growing interest in developing session-based resources in recent years.
Across these dimensions, existing datasets reveal consistent gaps in class distribution, platform coverage, linguistic diversity, modality, and data structure, collectively constraining model generalizability and pointing to clear priorities for future resource construction.

\subsection{Session-Based Datasets}\label{subsec:session dataset}
While the majority of existing cyberbullying datasets are constructed at the sentence level, session-based datasets have attracted increasing attention due to their ability to capture richer contextual and interactional information. In particular, cyberbullying often manifests as repetitive and evolving behaviors within conversations rather than isolated instances, and session-based data enable the modeling of such dynamics, including the progression and escalation of harmful interactions over time\cite{Yi2023LearningLH,Yi2022SessionbasedCD}. Moreover, compared with non-session-based data, session-based datasets more closely reflect real-world social media environments, where user interactions unfold through threads of posts and replies. This makes them more suitable for studying the lifecycle of cyberbullying and for supporting practical tasks such as detection, monitoring, and intervention.
In this section, we provide an overview of existing session-based datasets and briefly introduce their main characteristics.

\textbf{MySpace} \cite{d1}. One of the earliest and representative session-based cyberbullying datasets was constructed from MySpace forum threads. The dataset is organized as multi-user discussion threads, where each post is treated as a basic unit of analysis. To incorporate contextual information, the authors employ a sliding window over consecutive posts, with each window serving as the unit for annotation. Each window is labeled for the presence of cyberbullying based on majority agreement among multiple annotators. Each post consists of the user profile,  date, and content. This dataset represents one of the early attempts to model cyberbullying in conversational contexts rather than isolated messages.

\textbf{Instagram} \cite{d3}. This dataset is constructed from the Instagram platform, where each session is defined as a media post along with its associated comments. The data are organized around a central post and subsequent user interactions, forming content-centric conversational threads. Each session is accompanied by rich metadata, including captions, timestamps, and engagement signals such as likes and shares. To ensure sufficient contextual information, only sessions containing at least 15 comments are retained. Each session is annotated by five qualified annotators following detailed guidelines, and final labels are determined through aggregation. This dataset has been widely used for studying cyberbullying in social media environments with rich interaction contexts.

\textbf{Vine} \cite{d4}. This dataset is collected from the Vine platform, a video-based social network where users share short videos accompanied by user comments. Similar to the Instagram dataset, each session is defined as a media post together with its associated comments, forming content-centric conversational threads. The key distinction lies in the type of media, as sessions are initiated by videos rather than static posts. The dataset is constructed and annotated following the same procedure as the Instagram dataset, including the use of multiple annotators and aggregated labeling.

\textbf{SCCD} \cite{d19}. This dataset represents the first publicly available Chinese session-based cyberbullying dataset, constructed from the Weibo platform. Compared with earlier datasets, it exhibits a relatively balanced class distribution and reflects more recent social media environments. A key feature of this dataset lies in its fine-grained annotation at the comment level within each session. Specifically, individual comments are annotated along multiple dimensions, including expression patterns, sarcasm, and whether the content targets specific individuals or groups. Given the large volume of comments in each session, the dataset incorporates large language models within a human-in-the-loop framework to assist the annotation process. Due to the high cost associated with such fine-grained annotation, the overall dataset size remains relatively limited compared with other session-based datasets. Despite its relatively small scale, the dataset provides valuable support for in-depth analysis of cyberbullying behaviors in session-based contexts.

\textbf{CYBY23} \cite{d13}. This dataset is a publicly available session-based cyberbullying corpus constructed from Twitter threads, where each session consists of a main post and its associated replies. It introduces fine-grained annotation of bystander roles. In addition to labeling the aggression level of posts, the dataset further characterizes participant roles in replies, such as instigators, defenders, and neutral bystanders. Moreover, the annotation is conducted at multiple levels, including post-level, reply-level, and session-level, enabling a more comprehensive representation of interaction dynamics. Due to the complexity of such multi-level and multi-dimensional annotation, the dataset remains relatively limited in scale. Nevertheless, it provides valuable support for analyzing participant roles and behavioral dynamics in session-based cyberbullying.

\textbf{CYBY24} \cite{d20}. Building upon CYBY23, the authors further introduce CYBY24, a more comprehensive session-based dataset constructed from Twitter sessions. While maintaining a similar session structure, this dataset extends the annotation scheme by jointly annotating fine-grained cyberbullying severity and bystander roles. Specifically, main posts are labeled with fine-grained cyberbullying categories, while replies are annotated with corresponding participant roles. Compared with CYBY23, CYBY24 is constructed at a larger scale and adopts a more organized annotation structure, which is more conducive to analyzing the relationship between cyberbullying severity and participant behaviors. Together, these datasets reflect the growing focus on modeling participant roles and interaction dynamics in session-based cyberbullying research.

\textbf{PCCD} \cite{d21}. This dataset is a session-based cyberbullying corpus constructed from Twitter (X), consisting of tweets organized into conversation threads. Each session is formed by a sequence of tweets sharing a common thread identifier, enabling the modeling of interaction contexts. A key characteristic of this dataset is its focus on code-mixed content, particularly Chinese–English (Cantonese), addressing the scarcity of mixed-language resources. In terms of annotation, the dataset incorporates abuser–victim identity labels and defines cyberbullying at the session level based on factors such as aggression, repetition, and power imbalance. Additionally, it adopts a privacy-aware design by anonymizing user-related information. Compared with earlier datasets, PCCD provides richer annotations for modeling participant relationships and interaction dynamics in session-based cyberbullying.

In addition to the aforementioned datasets, recent studies have further extended existing session-based resources with more fine-grained annotations. For instance, \citet{d22} extracted a subset of sessions from the Instagram dataset and manually annotated each comment to capture temporal properties, enabling the analysis of how interactions evolve over time. Building upon this, \citet{d23} further enriched the annotation scheme by introducing diverse fine-grained attributes, such as the intent behind comments, to facilitate a deeper exploration of content patterns within sessions.

Taken together with datasets such as SCCD and CYBY, these efforts reveal a clear and consistent trend in session-based cyberbullying research. Specifically, dataset construction is gradually evolving from coarse-grained labeling toward more structured, fine-grained, and interaction-aware annotation frameworks. This shift enables a more comprehensive understanding of the dynamics of cyberbullying, including how harmful behaviors emerge, propagate, and are responded to within sessions.
Nevertheless, persistent challenges remain: most session-based datasets are limited in scale and confined to a single platform or language, fine-grained annotation schemes lack unified standards across datasets, and multimodal session data remain scarce. Addressing these gaps—through larger-scale collection, cross-platform design, and standardized annotation protocols—represents a critical direction for advancing session-based cyberbullying research.

\section{Challenges and Future Directions}
\label{sec:challenges}
Despite substantial advancements in detection, modeling, early warning, and intervention, existing methodologies still struggle to meet the demands of highly dynamic and complex governance in real-world social media scenarios. 
In this section, we summarize key challenges and future directions from four perspectives.

\subsection{From Detection to Context and Lifecycle-Aware Governance}
\label{subsec:tolifecycle}

A fundamental challenge in cyberbullying research lies in the misalignment between its inherently social-interactive nature and prevailing computational paradigms. 
Although cyberbullying is typically characterized by repetition, hostility, and power imbalance, most existing approaches reduce it to an isolated, post-level harmful content detection task, thereby overlooking its temporal, relational, and dynamic evolutionary characteristics \cite{kim2021human}. 
In practice, cyberbullying typically emerges from sustained interactions and the escalation of conflicts \cite{thomas2021sok}.
Recent research into session-level detection has begun to emphasize the importance of capturing contextual information \cite{yi2023session}. 
A vital future direction is to further integrate temporal signals, role transitions, escalation paths, and optimal intervention timing into a unified framework. 
Such developments would enable a more comprehensive, lifecycle-aware understanding of how cyberbullying emerges, propagates, and can be effectively mitigated \cite{milosevic2022artificial}.
 
\subsection{Explainability and Accountability in Moderation}
\label{subsec:explainability}

The efficacy of cyberbullying governance is severely constrained by the lack of explainability in automated systems. 
While current black-box models are capable of identifying harmful content, they often fail to provide justifications aligned with platform policies or legal standards.
This opacity not only erodes user trust but also complicates the appeal process due to a lack of transparency \cite{kiritchenko2021confronting, singhal2023sok}.
At the legal level, governance frameworks are standardizing globally, as evidenced by the European Union's \textit{Digital Services Act (DSA)} and China's \textit{Civil Code}, alongside the \textit{Guiding Opinions on Punishing Cyberbullying Crimes in Accordance with the Law} issued by the Supreme People's Court and other ministries. 
Consequently, explainable moderation is pivotal not only for platform-level violation judgments but also for the legal identification of illicit acts, evidence preservation, and the determination of liability. 
Future research must prioritize explainable moderation, human-in-the-loop review, and robust appeal mechanisms, fostering evidence-based governance and accountability workflows that bridge platform policies with judicial regulations.

\subsection{Bias, Fairness, and Low-Resource Governance}
\label{subsec:ethics}

Algorithmic bias, fairness, and adaptation to low-resource scenarios represent unavoidable hurdles for real-world deployment. 
Existing studies indicate that datasets and models for harmful language are often compromised by sampling bias, annotator subjectivity, and narrow task definitions, leading to systemic misjudgments across dialects, minority expressions, and specific cultural contexts \cite{balayn2021automatic}. 
In the context of cyberbullying, such biases may result in the over-censorship of benign speech or the under-protection of vulnerable users. 
These issues are particularly acute in multilingual and low-resource environments \cite{hee2024recent}. 
A key future direction is the establishment of evaluation systems that account for both fairness and contextual nuances, including sub-group error analysis, culturally-sensitive annotation protocols, and enhanced modeling support for low-resource languages and regional dialects.

\subsection{The Dual-Use of Large Language Models}
\label{subsec:llm_dual_use}

Large Language Models (LLMs) introduce a fundamental dual-use challenge in cyberbullying governance.
On one hand, they offer transformative potential in long-context understanding, multimodal moderation, and counter-narrative generation, enabling more granular identification and proactive intervention \cite{hee2024recent}. 
On the other hand, their generative capabilities can be exploited by malicious actors to produce more subtle, personalized, and adaptive forms of abusive content, such as paraphrased, sarcastic, or adversarially crafted attacks \cite{franco2025integrating}. 
This significantly weakens the effectiveness of existing moderation systems and exacerbates the adversarial nature of cyberbullying governance.

Beyond their role as moderation tools, LLM-based agent simulation opens new possibilities for governance by constructing controllable environments that model aggressors, victims, bystanders, and moderators. 
Such frameworks support analysis of conflict escalation, opinion radicalization, and bystander dynamics, and provide sandboxes for testing intervention strategies and optimizing platform policies before deployment. 
However, current agent simulation faces challenges including consensus bias, limited cross-cultural generalization, and the risk of misrepresenting vulnerable user groups.

Accordingly, future research should explore how to leverage LLMs to enhance governance performance while addressing their risks in adversarial settings. This includes developing policy-aware intervention strategies, fact-based generation control, more realistic multi-agent evaluation environments, and standardized protocols for assessing abuse, escalation, and adversarial evasion risks~\cite{bonaldi2024nlp}.


\section{Conclusion} 
\label{sec:conclusion}

Cyberbullying on social media is not merely a content detection problem, but a broader governance challenge spanning harmful content identification, behavioral interaction, diffusion, and response. 
In this paper, we propose a unified full-lifecycle framework and ground it in a systematic analysis of the literature across four interconnected stages:
content identification, user and behavior modeling, diffusion dynamics and early warning, and intervention and governance.
Our analysis shows that, despite substantial progress, current research remains fragmented across tasks and communities. Key challenges persist in context and lifecycle modeling, explainability and accountability, fairness and low-resource governance, and the dual-use risks of large language models.
Overall, this work highlights the need to move beyond isolated technical tasks toward more integrated, human-centered, and governance-aware approaches to cyberbullying on social media.


\bibliographystyle{ACM-Reference-Format}
\bibliography{main}

\appendix









\end{document}